\documentclass[pdflatex,sn-apa, iicol]{sn-jnl}

\usepackage{graphicx}%
\usepackage{multirow}%
\usepackage{amsmath,amssymb,amsfonts}%
\usepackage{amsthm}%
\usepackage{mathrsfs}%
\usepackage{xcolor}%
\usepackage{textcomp}%
\usepackage{manyfoot}%
\usepackage{algorithm}%
\usepackage{algorithmicx}%
\usepackage{algpseudocode}%
\usepackage{listings}%
\usepackage{subfigure}%
\usepackage{rotating} %
\usepackage{placeins} %
\usepackage{caption}  %
\usepackage{subfigure}
\usepackage{longtable} %
\usepackage{colortbl}
\usepackage{array}
\usepackage{booktabs}%
\usepackage{hyperref}
\usepackage{balance}

\definecolor{myorange}{RGB}{191,80,0} %
\definecolor{mygreen}{RGB}{0,155,72} %
\definecolor{myblue}{RGB}{0,80,200} %

\newtheorem{lemma}{Lemma}

\theoremstyle{thmstyleone}%
\newtheorem{theorem}{Theorem}
\newtheorem{proposition}[theorem]{Proposition}%

\theoremstyle{thmstyletwo}%

\theoremstyle{thmstylethree}%

\begin{document}

\title[Article Title]{Semantic-guided Fine-tuning of Foundation Model for Long-tailed Visual Recognition}


\author[1]{\fnm{Yufei} \sur{Peng}}\email{csyfpeng@comp.hkbu.edu.hk}

\author[1]{\fnm{Yonggang} \sur{Zhang}}\email{csygzhang@comp.hkbu.edu.hk}

\author*[1]{\fnm{Yiu-ming} \sur{Cheung}}\email{ymc@comp.hkbu.edu.hk}

\affil[1]{\orgdiv{Department of Computer Science}, \orgname{Hong Kong Baptist University}, \orgaddress{\city{Hong Kong SAR}, \country{China}}}

\abstract{The variance in class-wise sample sizes within long-tailed scenarios often results in degraded performance in less frequent classes. Fortunately, foundation models, pre-trained on vast open-world datasets, demonstrate strong potential for this task due to their generalizable representation, which promotes the development of adaptive strategies on pre-trained models in long-tailed learning. Advanced fine-tuning methods typically adjust visual encoders while neglecting the semantics derived from the frozen text encoder, overlooking the visual and textual alignment. To strengthen this alignment, we propose a novel approach, \textbf{S}em\textbf{a}ntic-\textbf{g}uid\textbf{e}d fine-tuning of foundation model for long-tailed visual recognition (Sage), which incorporates semantic guidance derived from textual modality into the visual fine-tuning process. Specifically, we introduce an SG-Adapter that integrates class descriptions as semantic guidance to guide the fine-tuning of the visual encoder. The introduced guidance is passesed through the attention mechanism and enables the model to focus more on semantically relevant content, strengthening the alignment between the visual and textual modalities. Due to the inconsistent class-conditional distributions neglected by the existing loss function, the resulting prediction bias causes performance improvements for the tail class less than for the head class, even when the multi-modal alignment is enhanced. To address this challenge, we propose a novel distribution mismatch-aware compensation factor, which is specifically designed to rectify the prediction bias caused by the ignored inconsistent distribution based on our theoretical analysis, and is seamlessly integrated into the loss function. Extensive experiments on benchmark datasets demonstrate the effectiveness of the proposed Sage in enhancing performance in long-tailed learning.

}

\keywords{Long-tailed learning, Fine-tuning strategy, Foundation model, Imbalanced image classification}

\maketitle
\balance
\section{Introduction}
\begin{figure*}[t]
\centering
\includegraphics[width=6.2in]{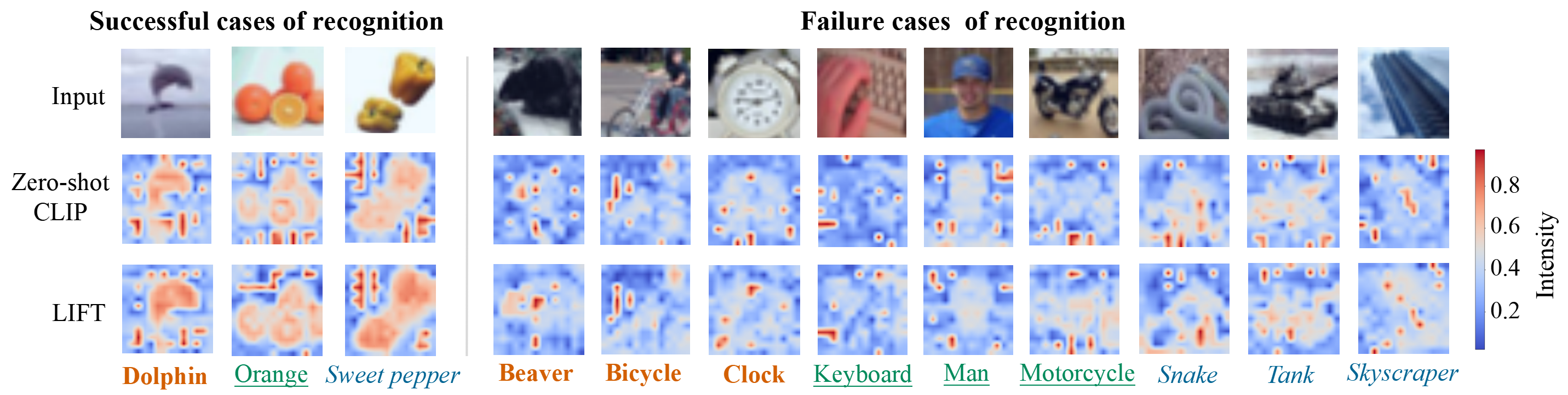}
\caption{An example of visual attention maps generated using different methods, where the head class label is represented in \textbf{\textcolor{myorange}{orange}}, the medium class label in \textcolor{mygreen}{\underline{green}}, and the tail class label in \textit{\textcolor{myblue}{blue}}, based on CIFAR-100-LT with $\beta=100$.}
\label{fig:atten_hmt}
\end{figure*}
Real-world data inherently follow an imbalanced distribution, which violates the assumption of balanced distribution in the majority of deep learning research~\citep{johnson2019survey, tang2020long}. The long-tailed distribution is a special but crucial case in imbalanced learning. In such scenarios, a few classes, referred to as head class, contain abundant samples, while the majority, known as tail class, has much fewer samples. This distribution presents significant challenges for deep learning models, particularly in achieving consistent performance across all classes when trained on highly imbalanced data. Due to the quantity difference, vanilla training procedures for deep neural networks tend to be dominated by the abundance of head samples, leading to insufficient training in the less frequent classes (i.e., classes with much fewer training samples than others) and resulting in degraded performance in these under-represented categories~\citep{yang2022survey, li2022gcl, li2022kps}. Tackling this imbalance is essential for developing more fair and effective models, especially in computer vision tasks.

Existing approaches for long-tailed visual recognition can be categorized into two main paradigms based on their underlying architectures: deep neural network-based methods and foundation model-based methods. Advanced techniques with deep neural networks primarily encompass logit adjustment strategies and two-stage frameworks. Logit adjustment methods assign class-wise margins to achieve unbiased loss functions, with margin assignment typically based on balanced error~\citep{menon2020la}, key point regularization~\citep{li2022kps}, or local Lipschitz continuity constraints~\citep{wang2023ddc}. Nevertheless, establishing a unified optimal margin allocation remains a challenge. Two-stage frameworks, on the other hand, have emerged as a predominant architecture in long-tailed learning. In the first stage, models are trained on imbalanced data, with representation learning improved through various augmentation strategies, ranging from the conventional techniques like Mixup~\citep{zhong2021mislas} to novel approaches such as noise injection~\citep{li2022gcl}. In the second stage, current studies incorporate advanced techniques, including label smoothing~\citep{zhong2021mislas}, feature map fusion~\citep{li2024h2t}, and feature uncertainty-based augmentations~\citep{ma2024fur} to facilitate unbiased classifier optimization. While these methods effectively leverage long-tailed data for robust feature extraction and employ rebalanced data for classifier balancing, they suffer from extended training requirements and challenges arising from distribution differences in training data between stages.

In contrast, foundation models offer a fundamentally different solution for long-tailed visual recognition. With the development of foundation models, researchers have increasingly realized the importance of the rich multi-modal common sense embedded within pre-trained models~\citep{yang2024mma, jin2024mv, khattak2023maple}. Regarding the issue of degraded performance in the less frequent classes caused by finite samples~\citep{yin2019feature}, foundation models pre-trained on vast amounts of data demonstrate significant potential. Existing methods~\citep{zhao2024ltgc, ma2022TACKLE} exploit foundation models to gather extensive web-sourced data, seeking to mitigate data quantity imbalances while extracting class-discriminative features. However, these techniques require additional storage capacity for the extra data and are vulnerable to noise generated during data collection. To address these challenges, some studies retrain the entire foundation model using long-tailed datasets~\citep{ma2021ballad, song2023ECVL}, while such approaches incur prohibitively high computational costs. Given the generalizable representations encoded by the vision-language foundation models, which make them particularly effective at handling under-represented classes in long-tailed image classification task, recent researches~\citep{shi2023lift} have shifted focus from retraining entire models to developing parameter-efficient adaptation techniques for vision-language pre-trained models on long-tailed data. These methods achieve notable performance enhancements with minimal additional parameters and fewer training epochs, offering a more practical and resource-efficient fine-tuning solution for long-tailed visual recognition.

Despite these advancements achieved with the help of vision-language foundation models, which align the visual and textual modalities, existing fine-tuning approaches for the visual modality predominantly rely on visual features alone. This reliance weakens the multi-modal alignment of the fine-tuned model. Consequently, neglecting semantics derived from the textual modality results in the model placing insufficient emphasis on semantically relevant content, as shown by the attention maps depicted in Fig.~\ref{fig:atten_hmt}. The attention maps are produced by zero-shot CLIP~\citep{radford2021clip} and the latest fine-tuning long-tailed learning method, LIFT~\citep{shi2023lift}, covering head, medium, and tail classes with labels tokenized in various formats. Compared with the successful cases of recognition, the failure cases reveal that the model tends to allocate attention to incorrect or redundant areas, or exhibits insufficient focus on semantically relevant content. Meanwhile, these visualizations highlight that erroneous attention allocation may happen across all classes, potentially impeding accurate recognition in long-tailed image classification. Therefore, incorporating semantic information during visual encoder fine-tuning is essential to strengthen the alignment between visual and textual modalities.


To alleviate the above issues, we propose a novel fine-tuning strategy, \textbf{S}em\textbf{a}ntic-\textbf{g}uid\textbf{e}d fine-tuning of foundation model (Sage), which incorporates semantic guidance into the fine-tuning process of the visual encoder of CLIP model for long-tailed visual recognition. Specifically, we integrate the textual description of each class into the proposed \textbf{S}emantic-\textbf{G}uide Adapter (SG-Adapter) when fine-tuning the visual encoder. The introduced class-specific description fuses the semantics into the multi-head self-attention layer in each block within the visual encoder, and effectively guides the visual encoder to focus more on semantic-related visual areas. As a result, this approach reinforces the alignment between the visual and textual modalities. However, even with the enhanced alignment, the tail class exhibits substantially fewer performance improvements relative to the head class. A potential reason for this is that the commonly used balanced loss function overlooks the discrepancy of the class-conditional distribution between the imbalanced training set and the balanced test set in long-tailed tasks. The inconsistent class-conditional distribution is prone to happen in the less frequent classes with limited samples, resulting in the learning of partial distributions rather than capturing the complete class distribution for these classes~\citep{he2009learning}. Thus, incomplete estimation of the underlying distribution enlarges the class-conditional distribution difference, and neglecting this difference further induces prediction bias on the tail class, thereby constraining performance improvements. To mitigate the above issue, we build upon the existing loss function and theoretically derive a compensation factor, which reduces the prediction bias caused by the neglected inconsistent class-conditional distributions and effectively enhances performance on the tail class. We perform extensive experiments on widely used long-tailed benchmark datasets. The experimental results demonstrate that the proposed strategy, Sage, significantly and consistently outperforms previous methods.

The main contributions of this paper can be summarized as follows.
\begin{itemize}
    \item Motivated by our observation that the absence of textual modality during visual encoder fine-tuning tends to result in incorrect attention to semantically unrelated content across all classes, we propose a novel long-tailed fine-tuning strategy, termed Sage, which emphasizes strengthening the alignment between visual and textual modalities while adapting to the pre-trained model. The incorporation of textual features during the fine-tuning of the visual encoder surprisingly guides the model to focus more on semantic-related content.
    \item We propose SG-Adapter, which introduces class textual description as semantic guidance during the fine-tuning of the visual encoder. The introduced guidance is then passed through the multi-head self-attention layer within the visual encoder and leads the model to capture the semantic-related content. 
    \item To address the challenge where the performance of head class is improved more effectively than that of tail class, we theoretically analyze the biased prediction caused by the ignored inconsistent class-conditional distributions in existing loss function. Based on the analysis, we derive a distribution mismatch-aware compensation factor to alleviate the prediction bias and seamlessly integrate it into the loss function, promoting learning of less frequent classes.
\end{itemize}

\section{Notations and Preliminaries}
Following previous works~\citep{zhong2021mislas, li2022gcl, du2024proco, shi2023lift}, this paper mainly focuses on the long-tailed image classification task. Different from almost all of the existing methods, our work delves into fine-tuning techniques specifically tailored for long-tailed challenges. We follow the backbone architecture used in LIFT~\citep{shi2023lift}, which leverages pre-trained CLIP model~\citep{radford2021clip}, employing Transformers as the core architecture for both the visual and text encoders. In this section, we make an overview of the foundational concepts of CLIP model and the related AdaptFormer~\citep{chen2022adaptformer}.

\textbf{Notations.}
Let $(\mathbf{x},y)$ denote a sample where $\mathbf{x} \in \mathcal{X}$ and $y \in \mathcal{Y}$, with $|\mathcal{Y}| = C$. The sample $\mathbf{x}$ belongs to class $y$ and is drawn from a training set $\mathcal{D} =\{(\mathbf{x},y)\}$ consisting of $C$ classes. The sample sizes for each class are denoted as $N=\{n_1,n_2,....n_C\}$, where $n_i$ indicates the number of samples in the $i$-th class. Without loss of generality, we assume $n_1 \ge n_2 \ge ... \ge n_C$ in the context of long-tailed image classification. The imbalanced ratio of a long-tailed dataset is defined as $\beta=\frac{n_1}{n_C}$, where $n_1$ and $n_C$ is the sample size of the most and the least frequent class, respectively. We consider a classification model 
 $\mathcal{M}= \mathbf{W} \circ f_\theta$, where $f_\theta$ is the feature extractor, and $\mathbf{W}=\{\mathbf{w}_1,\mathbf{w}_2,...,\mathbf{w}_C\}$ refers to the weights of the classifier. 

Building on previous works~\citep{deng2019arcface,li2022gcl,shi2023lift}, we adopt a cosine similarity-based classifier for prediction. That is, the predicted logit of the $i$-th class for a given input $\mathbf{x}$ is obtained through $z_i = \mathbf{w}_i f_\theta(\mathbf{x})^\top$, where both the representation $f_{\theta}(\mathbf{x})$ and the classifier weight for the $i$-th class $\mathbf{w}_i$ are normalized with $f_{\theta}(\mathbf{x}) = \frac{f_{\theta}(\mathbf{x})}{\|f_{\theta}(\mathbf{x})\|_2}$ and $\mathbf{w}_i = \frac{\mathbf{w}_i}{\|\mathbf{w}_i\|_2}$. The commonly used balanced loss function $\ell(\cdot)$, LA~\citep{menon2020la}, is computed as follows:
\begin{equation}
\begin{split}
    \ell(\mathbf{x},y=i) = - \log \left(\frac{e^{{z}_i} \cdot n_i }{\sum_{k \in [C]} e^{{z}_k} \cdot n_k  }\right).
\end{split}
\label{eq:la}
\end{equation}
The balanced loss effectively alleviates the long-tailed issue to a certain extent.

\textbf{CLIP} is a large vision-language model designed to connect the visual and textual modalities. It consists of a text branch with a text encoder $\mathcal{T}$ and a vision branch with an visual encoder $\mathcal{V}$, and both are jointly pre-tained with contrastive objective on web-scale image-text pairs~\citep{yang2024mma}. In this case, the visual encoder $\mathcal{V}$ acts as a feature extractor. Given an image $\mathbf{x}$, it is initially divided into non-overlapping patches and prepended with a special \texttt{[CLS]} token. The concatenated features $\mathbf{f}\in\mathbb{R}^{b\times d}$ are then sequentially fed into $L$ Transformer blocks, where $b$ denotes the batch size and $d$ represents the dimension of the hidden layer, and we use $\mathbf{f}^{(l)}$ refer as the output feature of the $l$-th Transformer block. Each Transformer block contains a Multi-head Self-Attention layer (MSA) and a MLP layer~\citep{chen2022adaptformer}. In MSA of the $l$-th Transformer block, the output from the $(l-1)$-th Transformer block, denoted as $\mathbf{f}^{(l-1)}$, is linearly projected into three vectors for the $j$-th head, i.e., the query ($\mathbf{Q}^{j,(l)}$), key ($\mathbf{K}^{j,(l)}$), and value ($\mathbf{V}^{j,(l)}$). The $j$-th head is calculated by:
\begin{equation}
\label{eq:msa_eachhead}
\begin{split}
\operatorname{head}^{(l)}_j &= \operatorname{Attention}(\mathbf{Q}^{j,(l)}, \mathbf{K}^{j,(l)}, \mathbf{V}^{j,(l)}) \\
&=\operatorname{Softmax}\left(\frac{\mathbf{Q}^{j,(l)} {\mathbf{K}^{j,(l)}}^{\top}}{\sqrt{d_k}}\right) \mathbf{V}^{j,(l)},\\
\mathbf{Q}^{j,(l)} &= \mathbf{f}^{(l-1)}\mathbf{W}_q^{j,(l)}, \mathbf{K}^{j,(l)}= \mathbf{f}^{(l-1)}\mathbf{W}_k^{j,(l)}, \\
\mathbf{V}^{j,(l)}&= \mathbf{f}^{(l-1)}\mathbf{W}_v^{j,(l)}.
\end{split}
\end{equation}
Thus, the self-attention is then formulated as:
\begin{equation}
\label{eq:msa}
\begin{split}
\mathbf{\tilde{f}}^{(l)}&=\operatorname{MSA}^{(l)}\left(\mathbf{f}^{(l-1)}\right)\\
&= \operatorname{Concat}(\operatorname{head}^{(l)}_1, ..., \operatorname{head}^{(l)}_h)\cdot \mathbf{W}^{(l)}_o.
\end{split}
\end{equation}
Here, $h$ denotes the number of heads in a MSA, and $d_k$ represents the dimensionality of each head, with $d = h \cdot d_k$. In the $j$-th head of MSA, $\mathbf{W}_q^{j,(l)}\in \mathbb{R}^{d \times d_k}$, $\mathbf{W}_k^{j,(l)}\in \mathbb{R}^{d \times d_k}$, $\mathbf{W}_v^{j,(l)}\in \mathbb{R}^{d \times d_k}$, and $\mathbf{W}_o^{(l)}\in \mathbb{R}^{d \times d}$ are independent projection matrices, following~\citep{vaswani2017attention}. Combined with the original input feature of the $l$-th Transformer block $\mathbf{f}^{(l-1)}$, the obtained feature $\mathbf{\tilde{f}}^{(l)}$ are then sent into a LayerNorm layer (LN) and the MLP module (MLP), which is comprised of two fully connected layers and an GELU activation function. The operations performed within the $l$-th block can be summarized as:
\begin{equation}
\label{eq:total}
\begin{split}
\left\{\begin{array}{l} \mathbf{\tilde{f}}^{(l)}=\operatorname{MSA}^{(l)}\left(\mathbf{f}^{(l-1)}\right)+ \mathbf{f}^{(l-1)}, \\ 
\mathbf{f}^{(l)}=\operatorname{MLP}^{(l)}\left(\operatorname{LN}\left(\mathbf{\tilde{f}}^{(l)}\right)\right)+\mathbf{\tilde{f}}^{(l)}.\end{array}\right.
\end{split}
\end{equation}
The final output feature $\mathbf{f}^{(l)}$ is then sent into the $(l+1)$-th Transformer block.

\textbf{AdaptFormer} replaces the MLP module in the Transformer block with \textit{AdaptMLP}~\citep{chen2022adaptformer}. An AdaptMLP layer consists of a down-projection matrix $\mathbf{W}_{\text{down}}\in\mathbb{R}^{d\times r}$, a non-linear ReLU activation function $\sigma(\cdot)$, and an up-projection matrix $\mathbf{W}_{\text{up}}\in\mathbb{R}^{r\times d}$. Here, $r$ represents the bottleneck dimension. For a specific input feature $\mathbf{\tilde{f}}^{(l)}$ to the adapter, the operation in AdaptMLP can be summarized as follows~\citep{li2024adapterx}:
\begin{equation}
\label{eq:adaptformer}
\begin{split}
\mathbf{f}^{\prime (l)}&=\operatorname{Adapter}(\mathbf{\tilde{f}}^{(l)})\\
&=\sigma\left(\operatorname{LN}\left(\mathbf{\tilde{f}}^{(l)}\right) \cdot \mathbf{W}_{\text{down}} \right) \cdot \mathbf{W}_{\text {up}}.
\end{split}
\end{equation}
Different from Eq.~(\ref{eq:total}), the output features $\mathbf{f}^{(l)}$ in AdaptFormer obtained by residual connection, and the operations performed within the $l$-th block are currently summarized as:
\begin{equation}
\label{eq:output}
\begin{split}
\left\{\begin{array}{l} \mathbf{\tilde{f}}^{(l)}=\operatorname{MSA}^{(l)}\left(\mathbf{f}^{(l-1)}\right)+ \mathbf{f}^{(l-1)}, \\ 
\mathbf{f}^{\prime (l)}=\operatorname{Adapter}(\mathbf{\tilde{f}}^{(l)})\\
\mathbf{f}^{(l)}=\operatorname{MLP}^{(l)}\left(\operatorname{LN}\left(\mathbf{\tilde{f}}^{(l)}\right)\right)+s^{(l)} \cdot \mathbf{f}^{\prime (l)} + \mathbf{\tilde{f}}^{(l)},
\end{array}\right.
\end{split}
\end{equation}
where $s^{(l)}$ is a learnable parameter for combination at the $l$-th block. In this way, AdaptFormer achieves significant performance improvements by fine-tuning a limited number of parameters on a pre-trained model. This approach is particularly beneficial for the less frequent classes which suffer from insufficient data in long-tailed learning, as it leverages the powerful and generalizable representations learned from the extensive open-world data of the pre-trained model.

\section{Methodology}
In this section, we present a comprehensive explanation of the proposed Sage framework. We begin by outlining the underlying motivation (Section~\ref{sec:motivation}) and thoroughly discuss the key components in Sage, including the SG-Adapter (Section~\ref{sec:sg-adapter}), the compensation factor (Section~\ref{sec:loss}), and a feature interchange technique designed to further preserve the generality of the foundation model (Section~\ref{sec:recall}). Finally, we conclude with an overview of the implementation of the proposed method (Section~\ref{sec:method_overview}).

\subsection{Motivation}\label{sec:motivation}
The multi-modal foundation model CLIP aligns visual and textual modalities, and enables efficient adaptation for various downstream tasks. Previous fine-tuning-based long-tailed methods~\citep{dong2022lpt, shi2023lift} predominantly focus on adapting the visual encoder for image classification, exclusively relying on visual features. However, these approaches tend to overlook the semantic information derived from the textual modality, thereby weakening the alignment between the visual and textual modalities, causing the model to misplace its attention on the input image content. Our observations of visual attention maps show that fine-tuning the visual encoder relies solely on visual features causes LIFT to miss the critical image content. The mismatch between the visual areas and the class labels results in inaccurate class predictions, which ultimately degrades overall classification performance. This highlights the significance of incorporating semantic information during fine-tuning of the vision encoder $\mathcal{V}$. To this end, we propose the Semantic-Guided Adapter (SG-Adapter), which embeds class-related description as semantic guidance into the fine-tuning process of the visual encoder. The introduced guidance feeds semantics into the MSA layer, helping the model to focus on semantically relevant content, and thus enhancing the alignment between the visual and textual modalities. The detailed statement of the proposed SG-Adapter is provided in Section~\ref{sec:sg-adapter}.

The proposed SG-Adapter strengthens the multi-modal alignment, hence improving the overall classification performance. However, as the ablation studies presented in Section~\ref{sec:exp_abla_result}, the performance boost produced by SG-Adapter for the tail class is comparatively modest relative to that of the head class. A significant contributing factor is the inconsistency in class-conditional distribution between training and test data in long-tailed image classification. In other words, for a sample $\mathbf{x}$ of the $y$-th class, traditional loss functions follow the assumption of $P_s(\mathbf{x}\mid y)=P_t(\mathbf{x}\mid y)$, where the subscripts $s$ and $t$ mean the training and test data distributions, respectively. This assumption may break down in long-tailed scenarios, as the model tends to concentrate on learning the partial distribution of the subconcepts of the tail class rather than capturing its complete distribution, resulting in an inadequate estimation of the class-conditional distribution~\citep{he2009learning}. Therefore, the underestimated distribution enlarges the class-conditional difference between training and test data, and the ignorance of this difference leads to prediction bias in the tail class, resulting in minimal performance improvement. To address the issue, we first theoretically analyze the underlying causes of prediction bias arising from the commonly adopted balanced loss function, LA~\citep{menon2020la}. Building upon this analysis, we subsequently derive a distribution mismatch-aware compensation factor to mitigate biased prediction and improve performance on the less frequent classes, as discussed in Section~\ref{sec:loss}.

Using the SG-Adapter and the proposed compensation factor, Sage effectively leverages the advantages of the foundation model while enhancing multi-modal alignment. However, fine-tuning the foundation model to downstream long-tailed tasks may compromise its generality by introducing task-specificity for the downstream task. The generality of the foundation model provides a more comprehensive distribution for each class, complementing the limited distribution learned from the smaller downstream dataset. Thus, we propose adaptively preserving this generality with a feature interchange technique, allowing the visual and textual features from the model, before and after fine-tuning, can be crossed and recombined. Further details are provided in Section~\ref{sec:recall}.

\subsection{SG-Adapter}\label{sec:sg-adapter}
The proposed strategy, Sage, is built upon the CLIP model, which comprises both visual and text encoders. Following the previous discussion, incorporating semantics derived from textual modality during visual encoder fine-tuning is crucial to strengthen the alignment between the visual and textual modalities. This is achieved through the SG-Adapter, an enhanced adapter that introduces semantic guidance for the visual encoder.

\begin{figure*}[t]
\centering
\includegraphics[width=5.8in]{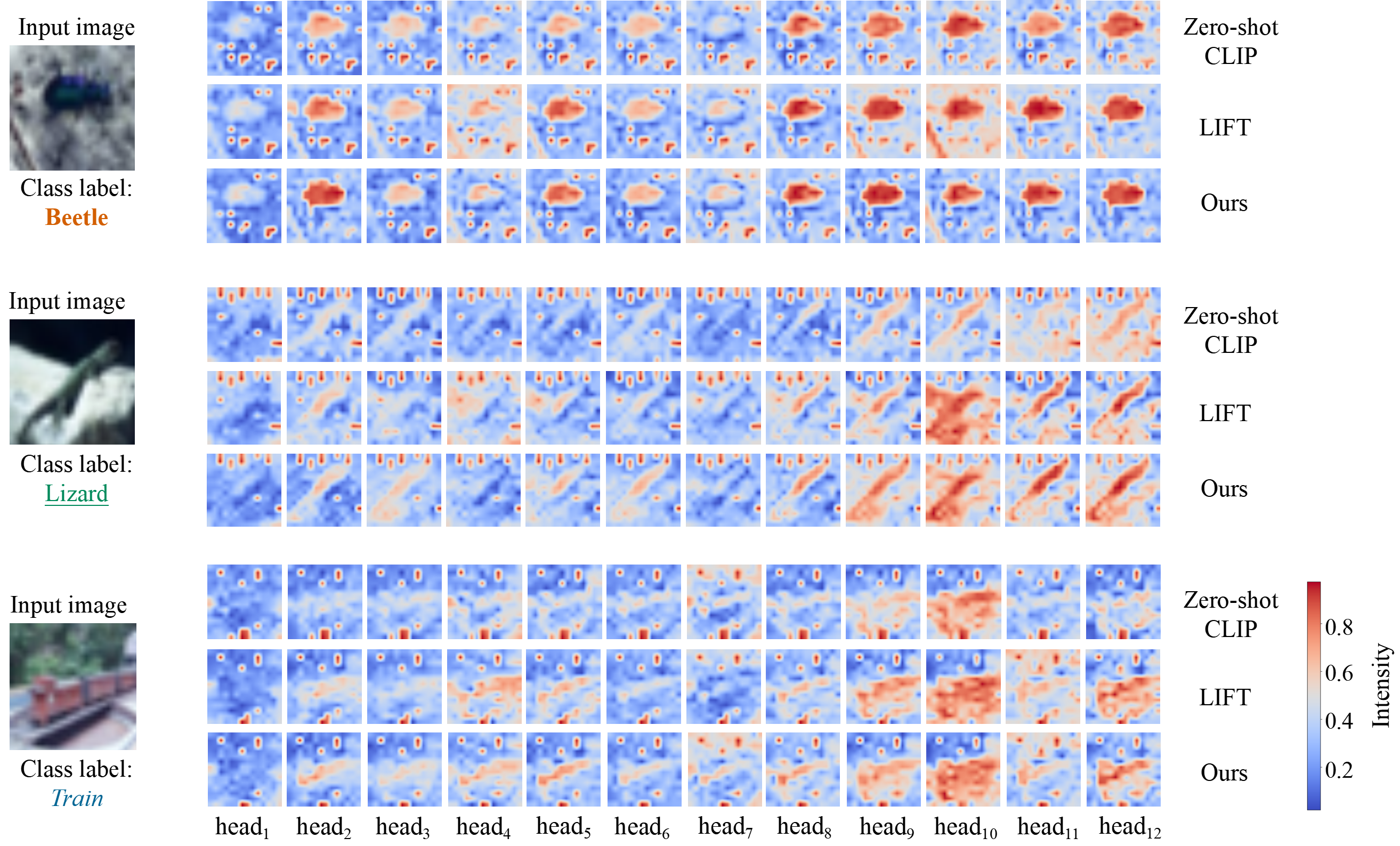}
\caption{Attention map visualization of $\{\operatorname{head}_j\}_{j=1}^{12}$, with the label for the head class is represented by \textbf{\textcolor{myorange}{orange}}, the medium class by \textcolor{mygreen}{\underline{green}}, and the tail class by \textit{\textcolor{myblue}{blue}}, based on CIFAR-100-LT with $\beta=100$.}
\label{fig:atten_multihead}
\end{figure*}
The semantic guidance for the visual encoder is derived from the classifier weights initialized with class textual descriptions. For each input class, we construct a set of textual templates $\{{\text{`a photo of a \texttt{[cls]}'}}, \dots,$ $\text{`an embroidered \texttt{[cls]}'}\}$, and `$\texttt{[cls]}$' is then replaced by the class name. These templates combined with the class names are fed into the frozen text encoder $\mathcal{T}$ of the zero-shot CLIP model to generate the template-specific textual embeddings $\{\mathbf{w}^{zs}_{i,j}\}_{j=1}^{T_n}$ for the $i$-th class, where $T_n$ denotes the total number of templates, and $\mathbf{w}^{zs}_{i,j}$ means the $j$-th template. The resulting textual features are averaged as $\mathbf{w}^{zs}_i=\frac{1}{T_n}\sum_{j=1}^{T_n} \mathbf{w}^{zs}_{i,j}$ to form $\mathbf{W}^{zs}=\{\mathbf{w}^{zs}_1, \mathbf{w}^{zs}_2, ..., \mathbf{w}^{zs}_C\}$, which initializes the trainable classifier weights $\mathbf{W}=\{\mathbf{w}_1, \mathbf{w}_2,..., \mathbf{w}_C\}$ for the corresponding classes. This process embeds class textual descriptions into the classifier weights $\mathbf{W}$, which serve as semantic guidance.

For each input image $\mathbf{x}$, it is sent into the visual encoder $\mathcal{V}$ with our proposed SG-Adapter, which compose an advanced visual encoder $\mathcal{V}^{\prime}$. The new visual encoder $\mathcal{V}^{\prime}$ keeps $L$ Transformer blocks, with each block combined with an SG-Adapter. Take the operation at the $l$-th block as an example, where $\mathbf{\tilde{f}}^{(l)}$ denotes the output feature from MSA and the residual connection as the same in Eq.~(\ref{eq:total}) at the $l$-th block. We first define three types of features used in the SG-Adapter,
\begin{equation}
\label{eq:define_feats}
\begin{split}
\mathbf{f}^{v} &= \operatorname{LN}\left(\mathbf{\tilde{f}}^{(l)}\right),\\ \mathbf{f}^{t} &= \operatorname{LN}\left(\operatorname{repeat}\left( {\bar{\mathbf{w}}}, b, 1\right)\right), \\
\mathbf{f}^{vt} &= \left(\mathbf{f}^{t}\cdot \mathbf{W}^{vt}_{\text{proj}}\right) * \mathbf{f}^{v},
\end{split}
\end{equation}
where the averaged classifier weight $\bar{\mathbf{w}}=\frac{1}{C}\sum_{i} \mathbf{w}_i$ with $\bar{\mathbf{w}}\in\mathbb{R}^{1\times d}$ is the introduced semantic guidance, and $*$ indicates element-wise multiplication. Also, we apply a modality projection matrix $\mathbf{W}^{vt}_{\text{proj}}\in\mathbb{R}^{d\times d}$ to project the semantic features into visual space. Define $\operatorname{repeat}\left(\mathbf{a}, b, i\right)$ as the operation that repeats $\mathbf{a}$ for $b$ times along the $i$-th dimension, leading the textual features $\mathbf{f}^{t}\in\mathbb{R}^{b\times d}$. Let $\mathbf{f}^{v}\in\mathbb{R}^{b\times d}$ and $\mathbf{f}^{vt}\in\mathbb{R}^{b\times d}$ represent the visual features and multi-modal features, respectively. With textual and multi-modal guidance, the core operations in SG-Adapter are as follows:
\begin{equation}
\label{eq:sg-adapter}
\begin{split}
\tilde{\mathbf{f}}^{vt} &= \left(\mathbf{f}^{v}\cdot \mathbf{W}^{v}_{\text{down}}\right) \| \\
&\hspace{0.5cm}\left(\mathbf{f}^{vt}\cdot \mathbf{W}^{vt}_{\text{down}} + s^{vt} \cdot\left( \mathbf{f}^{t}\cdot \mathbf{W}^{t}_{\text{down}}\right)\right),\\
\mathbf{f}^{\prime (l)}_{SG} &= \alpha \left(\sigma\left( \tilde{\mathbf{f}}^{vt} \right) \cdot \mathbf{W}^{vt}_{\text {up }}\right) + (1-\alpha)\left(\tilde{\mathbf{f}}^{vt} \cdot \mathbf{W}^{vt}_{\text {up }}\right),
\end{split}
\end{equation}
where $\|$ denotes concatenation along the feature dimension, resulting in $\tilde{\mathbf{f}}^{vt}\in \mathbb{R}^{b\times 2r}$. Similar to other adapters~\citep{chen2022adaptformer, yang2024mma}, we use down-projection matrices $\mathbf{W}^{v}_{\text{down}}, \mathbf{W}^{vt}_{\text{down}}, \mathbf{W}^{t}_{\text{down}}\in\mathbb{R}^{d\times r}$ and an up-projection matrix $\mathbf{W}^{vt}_{\text{up}}\in\mathbb{R}^{2r\times d}$. The learnable scale parameter $s^{vt}$ adaptively controls the contribution of the textual modality introduced, and $\alpha$ is a hyperparameter regulating the ratio of restoring negative activations. The operation in the $l$-th block, modified from Eq.~(\ref{eq:output}), is concluded as:
\begin{equation}
\label{eq:sg_summary}
\begin{split}
\left\{\begin{array}{l} \mathbf{\tilde{f}}^{(l)}=\operatorname{MSA}^{(l)}\left(\mathbf{f}^{(l-1)}\right)+ \mathbf{f}^{(l-1)}, \\ 
\mathbf{f}^{\prime (l)}_{SG}=\operatorname{SG-Adapter}\left(\mathbf{\tilde{f}}^{(l)}, \bar{\mathbf{w}}\right),\\
\mathbf{f}^{(l)}_{SG}=\operatorname{MLP}^{(l)}\left(\operatorname{LN}\left(\mathbf{\tilde{f}}^{(l)}\right)\right)+s^{(l)} \cdot \mathbf{f}^{\prime (l)}_{SG} + \mathbf{\tilde{f}}^{(l)}.
\end{array}\right.
\end{split}
\end{equation}
Here, the operations in $\text{SG-Adapter}\left(\cdot\right)$ are described in Eq.~(\ref{eq:define_feats}) and Eq.~(\ref{eq:sg-adapter}).

\textbf{Rationale Analysis.}
The key difference between SG-Adapter and previous long-tailed adaptation strategies is that we introduce multi-modal information to guide the visual encoder fine-tuning instead of relying solely on visual modality. We now explore how the semantic guidance helps model focus more on relevant content. 

\begin{figure*}[t]
\centering
\begin{minipage}{0.73\textwidth} 
    \centering
    \includegraphics[width=\textwidth]{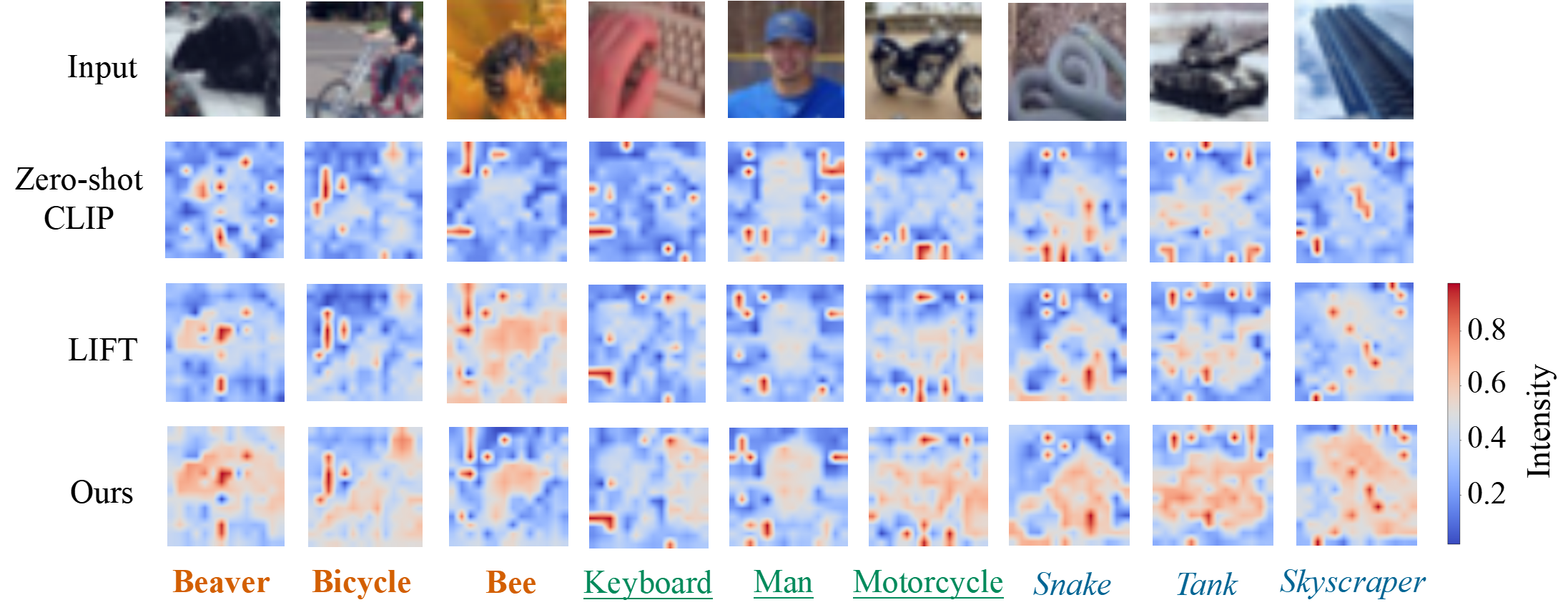}
\end{minipage}%
\hfill
\begin{minipage}{0.25\textwidth} 
    \caption{Visual attention maps generated using different methods, where the label for the head class is represented by \textbf{\textcolor{myorange}{orange}}, the medium class by \textcolor{mygreen}{\underline{green}}, and the tail class by \textit{\textcolor{myblue}{blue}}, based on CIFAR-100-LT with $\beta=100$.}
    \label{fig:atten_hmt_ours}
\end{minipage}
\end{figure*}
Given Eq.~(\ref{eq:sg_summary}), the output of the $l$-th block, $\mathbf{f}^{(l)}_{SG}$, is then passed as input to the MSA layer in the $(l+1)$-th block. For simplicity, we omit the superscript `$(l)$' and `$(l+1)$' for all parameters, and denote $\mathbf{f}^{(l)}_{SG}$ as $\dot{\mathbf{f}}$ in the following analysis. The conventional multi-head attention mechanism with $h$ heads in MSA, as defined in Eq.~(\ref{eq:msa_eachhead}) and Eq.~(\ref{eq:msa}), can be revisited as follows:
\begin{equation}
\label{eq:ori_multihead}
\begin{split}
\operatorname{MSA}(\dot{\mathbf{f}})& = \operatorname{Concat}(\operatorname{head}_1, ..., \operatorname{head}_h)\cdot \mathbf{W}_{o},\\
\operatorname{head}_j &= \operatorname{Attention}(\dot{\mathbf{f}}\mathbf{W}_{q}^j, \dot{\mathbf{f}}\mathbf{W}_k^j, \dot{\mathbf{f}}\mathbf{W}_v^j)\\ &=\operatorname{Softmax}\left(\frac{\mathbf{Q}^j {\mathbf{K}^j}^{\top}}{\sqrt{d_k}}\right) \mathbf{V}^j,\\
\mathbf{Q}^{j} = & \dot{\mathbf{f}}\mathbf{W}_q^{j}, \ \mathbf{K}^{j}= \dot{\mathbf{f}}\mathbf{W}_k^{j}, \ \mathbf{V}^{j}=\dot{\mathbf{f}}\mathbf{W}_v^{j}.
\end{split}
\end{equation}
For ease of analysis, we assume that the input $\dot{\mathbf{f}}$ of $\operatorname{head}_j$ can be represented as $\dot{\mathbf{f}}=[\mathbf{f}_1, \mathbf{f}_2]$, that is, $\dot{\mathbf{f}}$ is a combination of $\mathbf{f}_1$ and $\mathbf{f}_2$. Here, $\mathbf{f}_1\in\mathbb{R}^{b\times \frac{d}{2}}$ denotes the semantic related features while $\mathbf{f}_2\in\mathbb{R}^{b\times \frac{d}{2}}$ means the semantically unrelated features, that is, $\mathbf{f}_2 \perp \mathbf{f}_1$. Without loss of generality, we express $\mathbf{W}_{q}^j$ as $\begin{bmatrix} \dot{\mathbf{W}}_{q}^j \\ \ddot{\mathbf{W}}_{q}^j \end{bmatrix}$, with $\dot{\mathbf{W}}_{q}^j, \ddot{\mathbf{W}}_{q}^j \in \mathbb{R}^{\frac{d}{2}\times d_k}$. Thus, for $\mathbf{Q}^j$ in $\operatorname{head}_j$, we have:
\begin{equation}
\label{eq:head_i_q}
\begin{split}
\mathbf{Q}^j = \dot{\mathbf{f}}\cdot\mathbf{W}_{q}^j = [\mathbf{f}_1, \mathbf{f}_2] \begin{bmatrix} \dot{\mathbf{W}}_{q}^j \\ \ddot{\mathbf{W}}_{q}^j \end{bmatrix}=[\mathbf{f}_1\dot{\mathbf{W}}_{q}^j+\mathbf{f}_2\ddot{\mathbf{W}}_{q}^j].
\end{split}
\end{equation}
Similarly, for $\mathbf{K}^j$ and $\mathbf{V}^j$, we have:
\begin{equation}
\label{eq:head_i_kv}
\begin{split}
\quad\mathbf{K}^j=[\mathbf{f}_1\dot{\mathbf{W}}_{k}^j+\mathbf{f}_2\ddot{\mathbf{W}}_{k}^j],
\quad\mathbf{V}^j=[\mathbf{f}_1\dot{\mathbf{W}}_{v}^j+\mathbf{f}_2\ddot{\mathbf{W}}_{v}^j].
\end{split}
\end{equation}

Without considering $\operatorname{Softmax}$ and constant $\sqrt{d_k}$, Eq.~(\ref{eq:ori_multihead}) is calculated as:
\begin{equation}
\label{eq:head_i_qkv}
\begin{split}
&\left(\mathbf{Q}^j {\mathbf{K}^j}^{\top}\right) \mathbf{V}^j\\
&=\mathbf{f}_1\dot{\mathbf{W}}_{q}^j \dot{\mathbf{W}}_{k}^j\phantom{}^{\top} {\mathbf{f}_1}\phantom{}^{\top}{\mathbf{f}_1}\dot{\mathbf{W}}_{v}^j 
+ \mathbf{f}_1\dot{\mathbf{W}}_{q}^j \ddot{\mathbf{W}}_{k}^j\phantom{}^{\top} {\mathbf{f}_2}\phantom{}^{\top} {\mathbf{f}_1}\dot{\mathbf{W}}_{v}^j 
\\
&\hspace{0.5cm} + \mathbf{f}_2\ddot{\mathbf{W}}_{q}^j \dot{\mathbf{W}}_{k}^j\phantom{}^{\top} {\mathbf{f}_1}\phantom{}^{\top}{\mathbf{f}_1}\dot{\mathbf{W}}_{v}^j 
+ \mathbf{f}_2\ddot{\mathbf{W}}_{q}^j \ddot{\mathbf{W}}_{k}^j\phantom{}^{\top} {\mathbf{f}_2}\phantom{}^{\top}{\mathbf{f}_1}\dot{\mathbf{W}}_{v}^j \\
&\hspace{0.5cm} + \mathbf{f}_1\dot{\mathbf{W}}_{q}^j \dot{\mathbf{W}}_{k}^j\phantom{}^{\top} {\mathbf{f}_1}\phantom{}^{\top}{\mathbf{f}_2}\ddot{\mathbf{W}}_{v}^j
+ \mathbf{f}_1\dot{\mathbf{W}}_{q}^j \ddot{\mathbf{W}}_{k}^j\phantom{}^{\top} {\mathbf{f}_2}\phantom{}^{\top} {\mathbf{f}_2}\ddot{\mathbf{W}}_{v}^j
\\
&\hspace{0.5cm} + \mathbf{f}_2 \ddot{\mathbf{W}}_{q}^j \dot{\mathbf{W}}_{k}^j\phantom{}^{\top} {\mathbf{f}_1}\phantom{}^{\top}{\mathbf{f}_2} \ddot{\mathbf{W}}_{v}^j 
+ \mathbf{f}_2 \ddot{\mathbf{W}}_{q}^j \ddot{\mathbf{W}}_{k}^j\phantom{}^{\top} {\mathbf{f}_2}\phantom{}^{\top}{\mathbf{f}_2}\ddot{\mathbf{W}}_{v}^j.
\end{split}
\end{equation}
The derviation of Eq.~(\ref{eq:head_i_qkv}) is provied in Appendix~\ref{ap:multi_head}. In Eq.~(\ref{eq:head_i_qkv}), nearly all the addition terms are influenced by the semantic related features $\mathbf{f}_1$. Thus, the output of $\operatorname{head}_j$ exhibits a stronger correlation with the introduced semantics, and effectively filters out irrelevant content, as demonstrated by the comparative visualization in Fig.~\ref{fig:atten_multihead}. Suppose that in higher-dimensional space, we have $\operatorname{head}_j = [\operatorname{head}_j^t, \operatorname{head}_j^v]$ with $\operatorname{head}_j^t \perp \operatorname{head}_j^v$, meaning that $\operatorname{head}_j$ is composed of a semantically related part, $\operatorname{head}_j^t$ and a semantically unrelated part, $\operatorname{head}_j^v$. Expand this to all $\operatorname{head}$s in $\operatorname{MSA}(\cdot)$ as defined in Eq.~(\ref{eq:ori_multihead}) to obtain:
\begin{equation}
\label{eq:output_multihead}
\begin{split}
\operatorname{MSA}(\dot{\mathbf{f}}) =\operatorname{Concat}([\operatorname{head}_1^t, \operatorname{head}_1^v], ..., \\
[\operatorname{head}_h^t, \operatorname{head}_h^v])\cdot \mathbf{W}_{o},
\end{split}
\end{equation}
where each $\operatorname{head}_j$ contains a semantic guided feature $\operatorname{head}_j^t$. Consequently, incorporating semantics ensures that the output of the MSA is semantically related, thereby enhancing the alignment between the visual and textual modalities. This is further illustrated through a comparison between zero-shot CLIP, LIFT, and our method, as shown in Fig.~\ref{fig:atten_hmt_ours}. The visualization results show that compared with zero-shot CLIP and LIFT, the attention maps obtained by our method are more focused on the region of the object, while reducing the focus on irrelevant content.

\begin{figure*}[!t]
\centering
\includegraphics[width=5.7in]{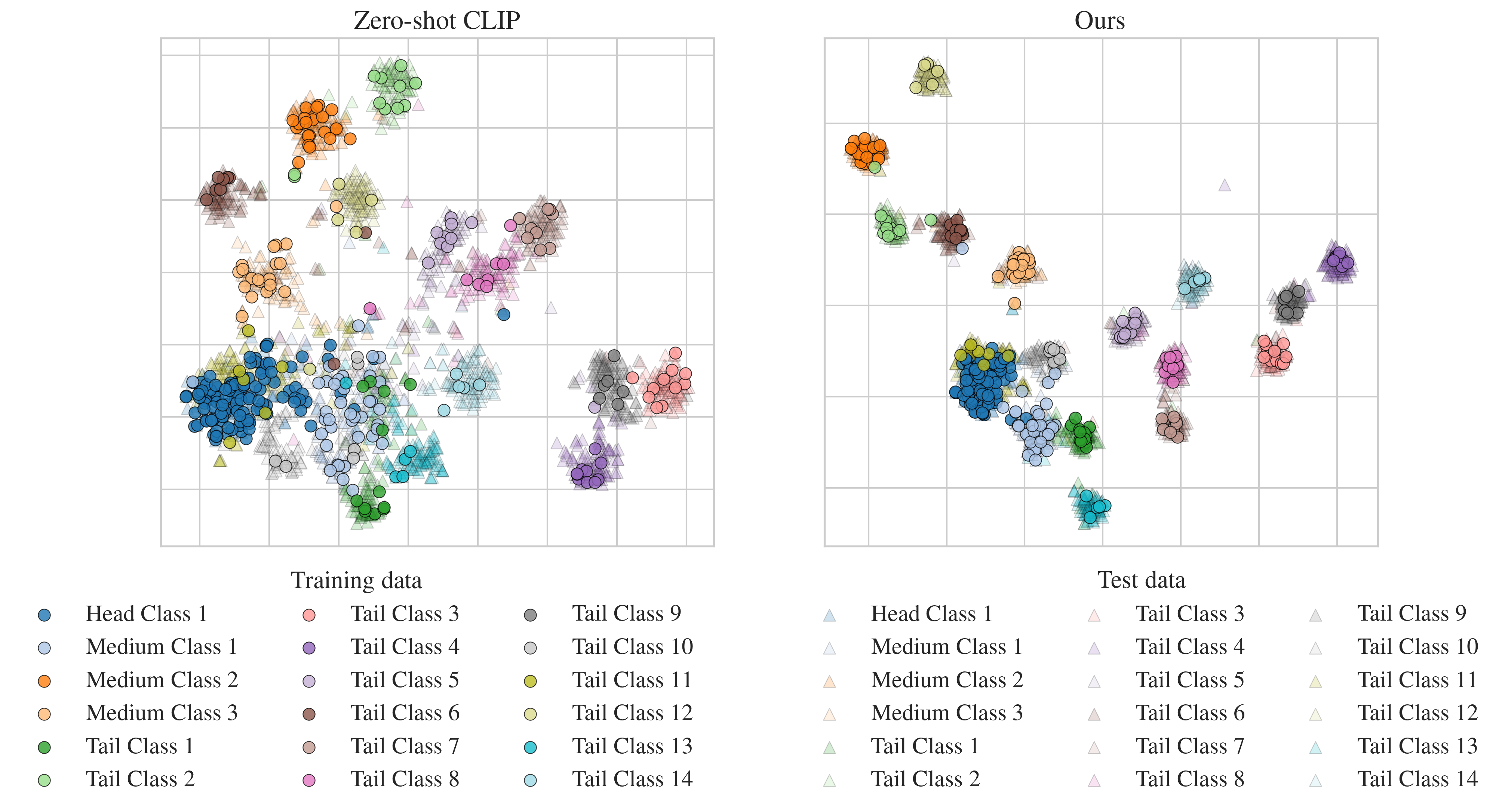}
\caption{t-SNE visualization of the embedding space for training and test data, where the embeddings are calculated with zero-shot CLIP and ours on the CIFAR-100-LT with $\beta=100$.}
\label{fig:pxy_zs_sgcf}
\end{figure*}
\subsection{Distribution mismatch-aware compensation factor}\label{sec:loss}
The proposed SG-Adapter greatly promotes the alignment between the visual and textual modalities, thereby leading the model to focus more on semantically related content. The resulting visual feature, denoted as $\mathbf{f}^{(L)}_{SG}$ and simplified as $\mathbf{f}$ hereafter, interacts with the classifier $\mathbf{W}$ to compute the prediction logit for the $i$-th class as $z_i = \mathbf{w}_i \mathbf{f}^\top$. However, the performance improvements for the tail class are noticeably smaller compared to those of the head class, as demonstrated in ablation studies in Section~\ref{sec:exp_abla_result}. For instance, the performance improvement observed in the head class is approximately 3.5 times greater than that of the tail class on the CIFAR-100-LT dataset with $\beta=50$. The disparity in performance gains between the tail and head classes necessitates a reconsideration of the existing balanced loss function for long-tailed tasks, which assumes a consistent class-conditional distribution between training and test data. In contrast to the common classification tasks with the consistent distributions of the training and testing sets, the differing distributions between the imbalanced training data and balanced test data in the long-tailed case potentially invalidate this assumption, and introduce additional challenges such as biased prediction. Therefore, it becomes imperative to address the limitations of existing loss functions arising from this inconsistent distribution.

In this section, we examine the widely adopted balanced loss, which assumes a consistent class-conditional distribution across all classes. However, this assumption neglects the practical challenge of incomplete estimation of the distribution for less frequent classes, resulting in biased predictions that hinder their performance improvement. To address this limitation, we theoretically derive a distribution mismatch-aware compensation factor that mitigates the prediction bias and further enhances learning for less frequent classes. This factor integrates seamlessly with the existing loss function. A comprehensive discussion follows.

\begin{lemma}[\cite{shi2023lift}]\label{lemma:1}
Underestimated class-conditional probability $P(\mathbf{x}\mid y=i)$ leads to an underestimated loss on the $i$-th class. Let $P_s$ and $P_t$ denote the probability distribution in the training and test domains, respectively. For the used loss function $\ell(\cdot)$, which is formulated as
\begin{equation}\notag
\label{eq:lift_loss}
\begin{split}
&\ell(\mathbf{x}, y = i) \\
&= - \log P_s(y = i \mid \mathbf{x})\\ &= - \log \left( \frac{e^{{z}_i + \log P_s(y = i) + \log \zeta_{s-t}(i)}}{\sum_{k \in [C]} e^{{z}_k + \log P_s(y = k) + \log \zeta_{s-t}(k)}} \right)\\
&= - \log \left( \frac{e^{{z}_i} \cdot n_i \cdot \zeta_{s-t}(i)}{\sum_{k \in [C]} e^{{z}_k} \cdot n_k \cdot \zeta_{s-t}(k)} \right),
\end{split}
\end{equation}
the assumption of $\{\zeta_{s-t}(i) = 1\}_{i=1}^{C}$ does not always hold in the long-tailed case. Here, $\zeta_{s-t}(i) = \frac{P_s(\mathbf{x} \mid y = i)}{P_t(\mathbf{x} \mid y = i)}$ for the $i$-th class. $P_s(y = i)$ represents the training label frequency of the $i$-th class, defined as $P_s(y = i)=\frac{n_i}{S_N}$, where $S_N=\sum_{j=1}^{C} n_j$.
\end{lemma}
Previous studies~\citep{menon2020la} assume that $\zeta_{s-t}(i)= 1$ for the $i$-th class, implying that the class-conditional distributions of the training and test data are consistent. However, as demonstrated in Lemma~\ref{lemma:1}, this assumption does not always hold in long-tailed learning due to discrepancies between the imbalanced training data and the balanced test data. That is, $P_s(\mathbf{x} \mid y = i)\neq P_t(\mathbf{x} \mid y = i)$, particularly for the less frequent classes, where the alignment between training and test data distributions is relatively weak caused by the insufficient learning on the complete distribution, as illustrated by the embeddings obtained via zero-shot CLIP in Fig.~\ref{fig:pxy_zs_sgcf}.

\begin{lemma}[Post-Compensation Strategy~\citep{hong2021lade}]\label{lemma:pc} 
For the input $\mathbf{x}$, the post-compensation strategy modifies the predicted logits of the $i$-th class $z_i$ as follows: 
\begin{equation}\notag
\label{eq:pc_strategy}
\begin{split}
z_i^{PC} = z_i - \log P_s(y = i) + \log P_t(y = i).
\end{split}
\end{equation}
Here, $z_i^{PC}$ denotes the post-compensated logit for the input $\mathbf{x}$ on the $i$-th class.
\end{lemma}

Lemma~\ref{lemma:pc} demonstrates that predictions can be effectively rectified by accounting for the label frequency discrepancies between training and test data. The existing balanced loss function given in Eq.~(\ref{eq:la}) assumes $\{\zeta_{s-t}(i) = 1\}_{i=1}^{C}$ and disregards the class-conditional distribution difference between the training and test data, leading to prediction bias according to our following analysis, which then results in limited performance gains for the less frequent classes. To overcome this, we build on Lemma~\ref{lemma:pc} and propose a distribution mismatch-aware compensation factor $\{\Lambda_i\}_{i=1}^C$ for all $C$ classes. This factor mitigates the biased prediction caused by the ignored inconsistent class-conditional distribution, thereby enhancing performance for the less frequent classes.
\begin{proposition}
The compensation factor $\{\Lambda_i = \mu n^{\gamma}_i \cdot \frac{S_{N}}{C \cdot n_{min}}\}_{i=1}^C$ modifies the loss function to reduce biased prediction caused by the neglected class-conditional distribution difference. The improved loss function, incorporating $\{\Lambda_i\}_{i=1}^C$, can be reformulated as:
\begin{equation}\notag
\begin{split}
&\ell(\mathbf{x}, y = i) \\
&= - \log \left( \frac{e^{z_i} \cdot n_i \cdot \left(\Lambda_i\cdot\zeta_{s-t}\left(i\right)\right)}{\sum_{k \in [C]} e^{z_k} \cdot n_k \cdot \left(\Lambda_k \cdot\zeta_{s-t}\left(k\right)\right)} \right)\\
&= - \log \left( \frac{e^{z_i} \cdot n_i \cdot \left(\mu n^{\gamma}_i \cdot\frac{S_{N}}{C\cdot n_{min}}\right)}{\sum_{k \in [C]} e^{z_k} \cdot n_k \cdot \left(\mu n^{\gamma}_k \cdot\frac{S_{N}}{C\cdot n_{min}}\right)} \right).
\end{split}
\end{equation}
\end{proposition}

\begin{figure*}[t]
\centering
\begin{minipage}{0.74\textwidth} 
    \centering
    \includegraphics[width=\textwidth]{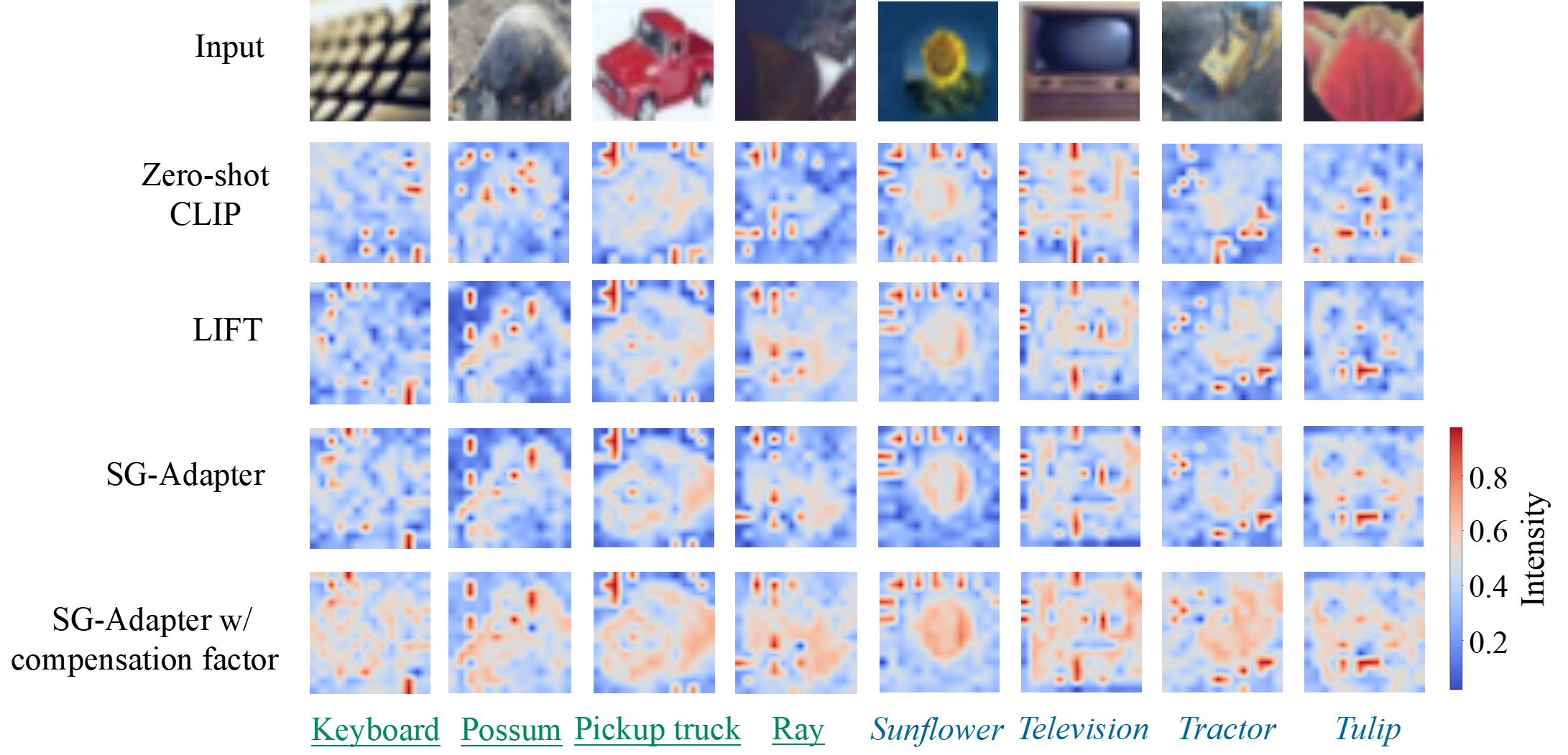}
\end{minipage}%
\hfill
\begin{minipage}{0.23\textwidth} 
    \caption{Visual attention maps generated using different methods, where the label for the medium class by \textcolor{mygreen}{\underline{green}}, and the tail class by \textit{\textcolor{myblue}{blue}}, based on CIFAR-100-LT with $\beta=100$.}
    \label{fig:atten_loss_hmt}
\end{minipage}
\end{figure*}
\textbf{Proof.} 
The theoretical analysis and derivation of the distribution mismatch-aware compensation factor $\Lambda_i$ for the $i$-th class are presented as follows. 

Define $\zeta_{s-t}^{\prime}(i) = \frac{P^{\prime}_s(\mathbf{x}\mid y=i)}{P_t(\mathbf{x}\mid y=i)}$, where $P^{\prime}_s(\mathbf{x}\mid y=i)$ represents the class-conditional distribution of the imbalanced training data for the $i$-th class. And $P_t(\mathbf{x}\mid y=i)$ indicates the class-conditional distribution for the balanced test data, as the same in $\zeta_{s-t}(i)$. In this case, $\zeta_{s-t}^{\prime}(i)$ is no longer always equal to $1$. Following the Bayes’ rule: $P(y=i \mid \mathbf{x}) = \frac{P(y=i) P(\mathbf{x} \mid y=i)}{P(\mathbf{x})}$, we transform $\zeta_{s-t}^{\prime}(i)$ into:
\begin{equation}
\label{eq:zeta_1}
\begin{split}
&\frac{P^{\prime}_s(\mathbf{x}\mid y=i)}{P_t(\mathbf{x}\mid y=i)}\\
&=\frac{P^{\prime}_s(y=i\mid \mathbf{x})\cdot P^{\prime}_s(\mathbf{x})}{P^{\prime}_s(y=i)}\cdot \frac{P_t(y=i)}{P_t(y=i\mid \mathbf{x})\cdot P_t(\mathbf{x})}\\
&= \frac{P^{\prime}_s(y=i\mid \mathbf{x})}{P_t(y=i\mid \mathbf{x})}\cdot \frac{P^{\prime}_s(\mathbf{x})}{P_t(\mathbf{x})}\cdot \frac{P_t(y=i)}{P^{\prime}_s(y=i)}.
\end{split}
\end{equation}
For simplicity, we refer to $P_s^{\prime}(y=i)$ and $P_t(y=i)$ as $P_s^{\prime}(i)$ and $P_t(i)$, respectively, throughout the following discussion. According to Lemma~\ref{lemma:pc}, the probability distribution on test data actually is $P_t(y=i\mid \mathbf{x})=\frac{e^{z^{PC}_i}}{\sum_{j \in [C]} e^{z^{PC}_j}}$, where $z^{PC}_i=z_i-\log P^{\prime}_s(i)+\log P_t(i)$ for the $i$-th class and $z_i$ denotes the underestimated logit during the training process. Define $\Theta(i) = \frac{P^{\prime}_s(y=i\mid \mathbf{x})}{P_t(y=i\mid \mathbf{x})}$ for the $i$-th class, and $\Theta(i)$ can be calculated as:
\begin{equation}
\label{eq:zeta_2}
\begin{split}
\Theta(i)&=\frac{\frac{e^{z_i}}{\sum_{j \in [C]} e^{z_j}}}{\frac{e^{z^{PC}_i}}{\sum_{j \in [C]} e^{z^{PC}_j}}}\\
&=\frac{P^{\prime}_s(i)}{P_t(i)}\cdot \frac{\sum_{j \in [C]} \frac{P_t(j)}{P^{\prime}_s(j)} e^{z_j}}{\sum_{j \in [C]} e^{z_j}}.
\end{split}
\end{equation}
Given Eq.~(\ref{eq:zeta_2}), in standard classification tasks where all classes have equal sample size in the training and test sets, namely, $\frac{P^{\prime}_s(i)}{P_t(i)}=\frac{P^{\prime}_s(j)}{P_t(j)}$ for any $i$-th and $j$-th class, resulting in $\Theta(i)=1$. However, in long-tailed scenarios, $\Theta(i)\neq1$ since $\frac{P^{\prime}_s(i)}{P_t(i)}\neq\frac{P^{\prime}_s(j)}{P_t(j)}$. This could be one of the detrimental factors exacerbating $\zeta_{s-t}^{\prime}(i) \neq 1$. In other words, the conventional loss function operates under the assumption of consistent class-conditional distributions, which largely overlooks the practical scenario where $\Theta(i) \neq 1$ in long-tailed learning.

However, the ignorance of $\Theta(i)\neq 1$ brings prediction bias. $\Theta(i)$ depicts the mapping from $P^{\prime}_s(y=i\mid \mathbf{x})$ to $P_t(y=i\mid \mathbf{x})$, that is, $\Theta(i)$ facilitates the transformation of the predicted logits from a long-tailed distribution to a balanced one. This enables the model to simulate training under a balanced distribution, even when trained on long-tailed data. Aligning training predictions with the corresponding test predictions, in other words, mapping training predictions to the corresponding predictions in a uniform distribution, potentially reduces the adverse effects of imbalanced sample size in long-tailed issues and further promotes optimization. Under the assumption of $\Theta = 1$, which may not hold in the long-tailed case, the head class, due to its significantly larger sample size in the training set compared to the test set, has a higher probability of correctly mapping its training predictions to corresponding test predictions. In contrast, the tail class often suffers from misalignment in the mapping due to insufficient representative training data. This misaligned mapping introduces bias, referred to as \textit{prediction bias} in this work, which causes training predictions to transfer into biased test predictions that deviate from those expected under a uniform distribution. This ultimately leads to suboptimal performance improvement for the tail class compared to the head class. Therefore, it is critical to mitigate this neglected prediction bias to further enhance the performance of the tail class, which suggests that we need to emphasize the inconsistent class-conditional distributions between training and test data during the training process.

Therefore, we incorporate a distribution mismatch-aware compensation factor into the existing loss function, aiming to explicitly consider the inconsistent class-conditional distributions during training. By combining Eq.~(\ref{eq:zeta_1}) and Eq.~(\ref{eq:zeta_2}), we derive $\zeta_{s-t}^{\prime}(i)$ as:
\begin{equation}
\label{eq:zeta_3}
\begin{split}
&\zeta_{s-t}^{\prime}(i) \\
&= \frac{P^{\prime}_s(i)}{P_t(i)}\cdot \Upsilon \cdot \frac{P^{\prime}_s(\mathbf{x})}{P_t(\mathbf{x})}\cdot \frac{P_t(i)}{P^{\prime}_s(i)}\\
&=\frac{P^{\prime}_s(\mathbf{x})}{P_t(\mathbf{x})}\cdot \Upsilon,
\end{split}
\end{equation}
where $\Upsilon=\frac{\sum_{j \in [C]} \frac{P_t(j)}{P^{\prime}_s(j)} e^{z_j}}{\sum_{j \in [C]} e^{z_j}}$. Building on the previous definition of $\zeta_{s-t}(i)=\frac{P_s(\mathbf{x} \mid y = i)}{P_t(\mathbf{x} \mid y = i)}=1$, which assumes that the class-conditional distributions of the training and test data are the same, the specific compensation factor $\Lambda_i$ for the $i$-th class can be calculated as follows: 
\begin{equation}
\label{eq:zeta_4}
\begin{split}
\Lambda_i &= \frac{\zeta_{s-t}^{\prime}(i)}{\zeta_{s-t}(i)}= \frac{P^{\prime}_s(\mathbf{x})}{P_t(\mathbf{x})}\cdot \Upsilon.
\end{split}
\end{equation}
Instead of directly calculating the value of $\Upsilon$, we derive its approximate upper bound under the conditions that $P_t(j)= \frac{1}{C}$, as balanced test data is used in long-tailed learning, and the training label frequency is $P_s^{\prime}(j)= \frac{n_j}{S_N}$ of the $i$-th class:
\begin{equation}
\Upsilon=\frac{1}{C}\cdot\frac{\sum_{j \in [C]} \frac{S_{N}}{n_j}\cdot e^{z_j}}{\sum_{j \in [C]} e^{z_j}}\leq \frac{1}{C}\cdot\frac{S_{N}}{n_{min}},
\end{equation}
where $n_{min}$ means sample size of the least frequent class. Thus, $\Lambda_i$ for the $i$-th class is obtained by:
\begin{equation}
\label{eq:zeta_5}
\begin{split}
\Lambda_i
&\leq \frac{P^{\prime}_s(\mathbf{x})}{P_t(\mathbf{x})} \cdot \frac{1}{C} \cdot \frac{S_{N}}{n_{min}}.
\end{split}
\end{equation}

\begin{figure*}[!t]
\centering
\includegraphics[width=5.8in]{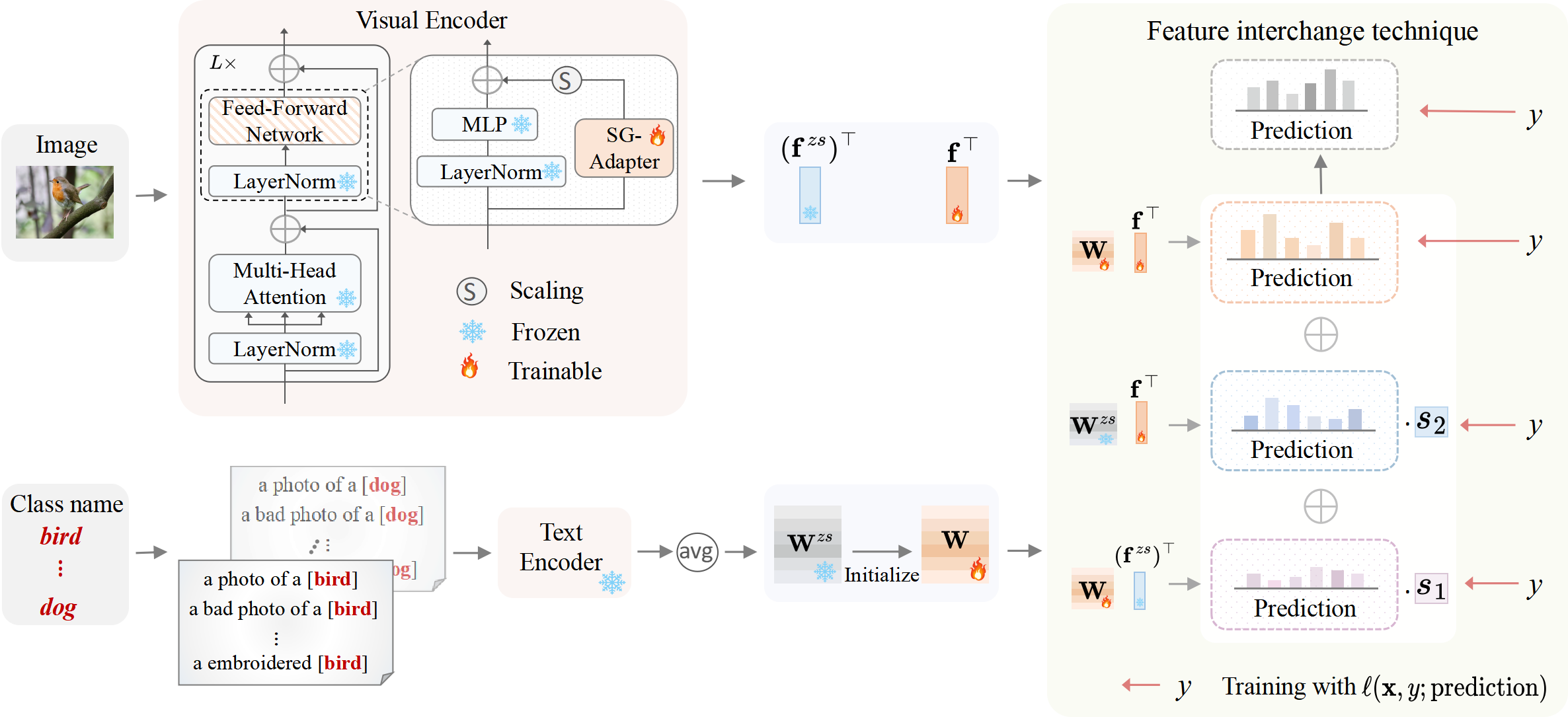}
\caption{The overall architecture of Sage, where \textbf{\textcolor{blue}{snow}} represents frozen parameters, and \textbf{\textcolor{orange}{fire}} represents trainable parameters.}
\label{fig:overview}
\end{figure*}
Performing accurate estimates $\frac{P^{\prime}_s(\mathbf{x})}{P_t(\mathbf{x})}=\frac{\sum_{k\in[C]} P^{\prime}_s(k) P^{\prime}_s(\mathbf{x}\mid y=k)}{\sum_{k\in[C]} P_t(k) P_t(\mathbf{x}\mid y=k)}$ is inherently challenging. Empirically, the value of 
$\frac{P^{\prime}_s(\mathbf{x})}{P_t(\mathbf{x})}$ tends to correlate with the sample size of each class in long-tailed scenarios (please refer to Appendix~\ref{ap:psx_ptx_ratio} for more analysis). Therefore, we approximate it as $\frac{P^{\prime}_s(\mathbf{x}_i)}{P_t(\mathbf{x}_i)}\approx \mu n^{\gamma}_i$ for the $i$-th class, where $\mathbf{x}_i$ represents sample $\mathbf{x}$ from the $i$-th class. Thus, the final form of the compensation factor $\Lambda_i$ for the $i$-th class is:
\begin{equation}
\label{eq:zeta_final}
\begin{split}
\Lambda_i \leq \mu n^{\gamma}_i \cdot\frac{S_{N}}{C\cdot n_{min}}.
\end{split}
\end{equation}
Let $\Lambda_i = \mu n^{\gamma}_i \cdot\frac{S_{N}}{C\cdot n_{min}}$ for improving computational efficiency, and the loss function $\ell(\cdot)$ combined with compensation factor now becomes:
\begin{equation}
\label{eq:lift_loss_final}
\begin{split}
&\ell(\mathbf{x}, y = i)\\
&= - \log \left( \frac{e^{z_i} \cdot n_i \cdot \left(\Lambda_i\cdot\zeta_{s-t}\left(i\right)\right)}{\sum_{k \in [C]} e^{z_k} \cdot n_k \cdot \left(\Lambda_k \cdot\zeta_{s-t}\left(k\right)\right)} \right)\\
&= - \log \left( \frac{e^{z_i} \cdot n_i \cdot \left(\mu n^{\gamma}_i \cdot\frac{S_{N}}{C\cdot n_{min}}\right)}{\sum_{k \in [C]} e^{z_k} \cdot n_k \cdot \left(\mu n^{\gamma}_k \cdot\frac{S_{N}}{C\cdot n_{min}}\right)} \right).
\end{split}
\end{equation}
\qed

This completes the proof. Applying the loss function defined in Eq.~(\ref{eq:lift_loss_final}), which highlights inconsistent class-conditional distribution, the model effectively reduces the prediction bias when mapping training predictions to the corresponding test predictions, especially for the tail class that is more affected. This improves performance for the less frequent classes and helps the model better focus on the correct visual areas within the samples from these classes, as illustrated in Fig.~\ref{fig:atten_loss_hmt}.

\subsection{Feature interchange technique}\label{sec:recall}
The proposed learning strategy, Sage, enhances the alignment between visual and textual modalities when adapting the foundation model to long-tailed image classification. Specifically, the SG Adapter (detailed in Section~\ref{sec:sg-adapter}) guides the model to focus more on semantically relevant content, while the distribution mismatch-aware compensation factor (elaborated in Section~\ref{sec:loss}) relieves performance improvement challenges for the less frequent classes. Both components strengthen multi-modal alignment across classes.

Rethinking the foundation model CLIP, which is trained on vast image-text paired data that far exceeds the scale of typical downstream benchmarks, possesses features of remarkable generality. Its visual and textual features, extracted through frozen encoders $\mathcal{V}$ and $\mathcal{T}$, are broadly applicable across diverse classes. In other words, with more abundant training data per class, foundation models trained with more data per class better capture comprehensive distributions, resulting in features that span a broader semantic space and richer visual concepts~\citep{bommasani2021opportunities, zhou2022coop}. However, this valuable generality diminishes when adapting the model to specific downstream tasks, as fine-tuning techniques often tailors these general features to become task-specific~\citep{Howard2018Universal, shen2022how}. Such task-specificity narrows the semantic space and limits the representation of diverse visual concepts, further restricting the ability of the model to generalize to unseen samples. Therefore, preserving the generality of features encoded by the foundation model is essential.

To address the aforementioned challenges, we propose FIT, a feature interchange technique that leverages features extracted from both pre-fine-tuning and fine-tuned models, namely, the features from the foundation model and the model after fine-tuning, to adaptively preserve generality. FIT is accomplished through the recomposition of the predicted logits, enabling alignment between pre-fine-tuning visual features and fine-tuned textual features, as well as between pre-fine-tuning textual features and fine-tuned visual features. Specifically, as previously described, $\mathbf{f}^{zs}$ represents the pre-fine-tuning visual feature derived from the zero-shot CLIP model, and $\mathbf{w}^{zs}_i$ is the classifier weight for the $i$-th class encoded by its text encoder, effectively serving as the corresponding pre-fine-tuning textual feature. Both $\mathbf{f}^{zs}$ and $\mathbf{w}^{zs}_i$ capture generality, as they are derived from the foundation model. Let $\mathbf{f}$ represent the fine-tuned visual feature extracted from the fine-tuned model, and $\mathbf{w}_i$ denote the fine-tuned textual feature derived from the classifier weight of the $i$-th class, where both keep the task-specificity for the downstream task. Therefore, the predicted logit ${\hat{z}}_i$ for the $i$-th class is then calculated as:
\begin{equation}
\begin{split}
{\hat{z}}_i &= {z}_i + s_1 z^{v}_i + s_2 z^{t}_i \\
&= \mathbf{w}_i \mathbf{f}^{\top} + s_1 \mathbf{w}_i \left(\mathbf{f}^{zs}\right)^{\top} + s_2 \mathbf{w}_i^{zs} \mathbf{f}^{\top},
\end{split}
\label{eq:final_logit}
\end{equation}
with ${z}_i=\mathbf{w}_i \mathbf{f}^{\top}$, $z^{v}_i = \mathbf{w}_i \left(\mathbf{f}^{zs}\right)^{\top}$, and $z^{t}_i = \mathbf{w}_i^{zs} \mathbf{f}^{\top}$. Here, $s_1$ and $s_2$ are learnable parameters designed to adaptively balance the generality and task-specificity of features. Let $\ell(\mathbf{x}, y = i; z_i)$ denotes the loss function associated with $z_i$ for a sample $\mathbf{x}$ on the $i$-th class, as defined in Eq.~(\ref{eq:lift_loss_final}). The overall loss function can now be expressed as:
\begin{equation}
\label{eq:loss_final_final}
\begin{split}
\ell^{\prime}(\mathbf{x}, y = i)= \ell(\mathbf{x}, y = i; z_i) + \lambda_1 \ell(\mathbf{x}, y = i; z^{v}_i)\\
+ \lambda_2 \ell(\mathbf{x}, y = i; z^{t}_i) + \lambda_3 \ell(\mathbf{x}, y = i; {\hat{z}}_i).
\end{split}
\end{equation}

In summary, the new logit $\hat{{z}}_i$ integrates the interchange between pre-fine-tuning features and fine-tuned features, effectively preserving the generality of the features obtained by the foundation model. This approach notably enhances performance for almost all classes, as demonstrated in ablation studies in Section~\ref{sec:exp_abla_result}.

\begin{algorithm}[t]
\caption{Sage}
\begin{algorithmic}[1]
\Require Training dataset $\mathcal{X}$, Test dataset $\mathcal{X}_{test}$, Batch size $b$, Learning rate $\eta$;
\Ensure Task-specific predictions;

\State \textbf{Initialization:} Initialize $\operatorname{SG-Adapter}$ parameters $\theta$ randomly. Initialize $\mathbf{W}$ with $\mathbf{W}^{zs}$;

\vspace{0.5em} 
\noindent \textbf{Training Phase:}
\vspace{0.3em}

\For {$\text{iter} = 1$ to $I_0$}
    \State Sample a batch size $b$ of samples $\mathcal{B}$ from $\mathcal{X}$;
    \State Extract visual features from the frozen visual encoder $\mathcal{V}$ with trainable $\operatorname{SG-Adapter}$, aligned with Eq.~(\ref{eq:sg_summary}); 
    \State Obtain the final predicted logits $\hat{\mathbf{z}}=\{\hat{z}_1,\hat{z}_2,..., \hat{z}_C \}$ through FIT given in Eq.~(\ref{eq:final_logit});
    \State Calculate loss $\ell^{\prime}(\mathbf{x}, y=i)$ as defined in Eq.~(\ref{eq:loss_final_final}), incorporating the compensation factor $\{\Lambda_i\}_{i=1}^C$ indicated in Eq.~(\ref{eq:lift_loss_final});
    \State Update parameters: $\Psi = \Psi - \eta \nabla_\Psi \mathcal{L}(\mathcal{B}; \Psi)$, where $\Psi = \{\theta, \mathbf{W}, \hat{s} \}$, and $\hat{s}=\{s_1,s_2\}$ in FIT.
\EndFor

\noindent \textbf{Inference Phase:}
\For {each test sample $\textbf{x} \in \mathcal{X}_{test}$}
    \State Extract visual features thorugh the frozen visual encoder $\mathcal{V}$ with trained $\operatorname{SG-Adapter}$, aligned with Eq.~(\ref{eq:sg_summary}); 
    \State  Produce the final predicted logits $\hat{\mathbf{z}}$ through FIT given in Eq.~(\ref{eq:final_logit}).
\EndFor

\end{algorithmic}
\label{al:algorithm}
\end{algorithm}

\subsection{Overview}\label{sec:method_overview}   
The overall architecture of Sage is illustrated in Fig.~\ref{fig:overview}, and is based on the CLIP model. Each input image $\mathbf{x}$ is first processed by an enhanced visual encoder $\mathcal{V}^{\prime}$, which incorporates a trainable SG-Adapter within each Transformer block. The output feature $\mathbf{f}$ is obtained from the $L$-th block of $\mathcal{V}^{\prime}$. Furthermore, let $\mathbf{f}^{zs}$ represent the features extracted from the $L$-th block of the frozen visual encoder $\mathcal{V}$ in the zero-shot CLIP model, excluding the proposed SG-Adapter.

For the textual branch, we construct a set of class-wise textual templates $\{\text{`a photo of a }$ $\text{\texttt{[cls]}'}, \dots, \text{`an embroidered \texttt{[cls]}'}\}$, and $\texttt{[cls]}$ is replaced with the class name. The $T_n$ templates are then fed into the frozen text encoder $\mathcal{T}$ of the zero-shot CLIP model to generate the textual features $\{\mathbf{w}^{zs}_{i,j}\}_{j=1}^{T_n}$ for each class, where $\mathbf{w}^{zs}_{i,j}$ denotes the $j$-th feature for the $i$-th class. The resulting textual features are then averaged as $\mathbf{w}^{zs}_i=\frac{1}{T_n}\sum_{j=1}^{T_n} \mathbf{w}^{zs}_{i,j}$, and initialize the trainable classifier weight $\mathbf{w}_i$ for the corresponding $i$-th class. In this way, the class textual descriptions are transmitted into the classifier weights $\mathbf{W}=\{\mathbf{w}_1, \mathbf{w}_2,..., \mathbf{w}_C\}$.

Next, the output visual feature $\mathbf{f}$ is acted with the classifier weights $\mathbf{W}$ to compute the predicted logits through a cosine classifier. To preserve the generality of the features encoded by the foundation model, we use FIT to interchange the features of the zero-shot CLIP model ($\mathbf{f}^{zs}$ and $\mathbf{W}^{zs}$) with those of the fine-tuned model ($\mathbf{f}$ and $\mathbf{W}$), and recombine them to produce the final predicted logits. The resulting predicted logit $\hat{z}_i$ for the $i$-th class is given as Eq.~(\ref{eq:final_logit}). The model is then trained using Eq.~(\ref{eq:loss_final_final}), where each term incorporates the widely adopted LA loss combined with the proposed distribution mismatch-aware compensation factor, as formulated in Eq.~(\ref{eq:lift_loss_final}). The compensation factor is designed to rectify the prediction bias that arises from the neglected inconsistent class-conditional distribution, alleviating the improvement challenges faced by the less frequent classes due to biased predictions. The overall procedure is summarized in Algorithm~\ref{al:algorithm}.

\section{Experiments}\label{sec:exp}
In this section, we present comprehensive experiments and analyses of the proposed method. We begin by introducing the datasets and implementation details in Section~\ref{sec:exp_dataset} and Section~\ref{sec:exp_implement}, respectively. Followed by comparison studies in Section~\ref{sec:exp_comparison} and ablation studies in Section~\ref{sec:exp_ablation}. Finally, we conclude the limitations in Section~\ref{sec:exp_limitations}.

\begin{table*}[t]
\centering
\setlength{\tabcolsep}{2pt} 
\caption{Comparison on CIFAR-100-LT with various imbalance ratios in terms of top-1 accuracy (\%). }
\begin{tabular}{l|c|c|c|ccc}
\hline
\multirow{2}{*}{\textbf{Methods}} & \multirow{2}{*}{\textbf{Backbone}} & \textbf{Learnable} & \multirow{2}{*}{\textbf{Epochs}} & \multicolumn{3}{c}{\textbf{Imbalance Ratio}} \\
& & \textbf{Params.} & & 100 & 50 & 10 \\
\hline
\multicolumn{7}{l}{\textbf{Training from scratch}} \\
\hline
LDAM~\citep{cao2019ldam} & ResNet-32 & 0.46M & 200 & 42.0 & 46.6 & 58.7 \\
BBN~\citep{zhou2019bbn} & ResNet-32 & 0.46M & 200 & 42.6 & 47.0 & 59.1 \\
CDB-W-CE~\citep{sinha2022cdb} & ResNet-32 & 0.46M & 200 & 42.6 & - & 58.7 \\
DiVE~\citep{he2021dive} & ResNet-32 & 0.46M & 200 & 45.4 & 51.1 & 62.0 \\
MiSLAS~\citep{zhong2021mislas} & ResNet-32 & 0.46M & 200+10 & 47.0 & 52.3 & 63.2 \\
BS~\citep{ren2020bs} & ResNet-32 & 0.46M & 400 & 50.8 & 54.2 & 63.0 \\
PaCo~\citep{cui2021paco} & ResNet-32 & 0.46M & 400 & 52.0 & 56.0 & 64.2 \\
BCL~\citep{zhu2022bcl} & ResNet-32 & 0.46M & 200 & 51.9 & 56.6 & 64.9 \\
FUR~\citep{ma2024fur} & ResNet-32 & 0.46M & 100+50+50 & 50.9 & 54.1 & 61.8 \\
ProCo~\citep{du2024proco} & ResNet-32 & 0.46M & 200 & 52.8 & 57.1 & 65.5 \\
DBM-NCL~\citep{son2024difficulty} & ResNet-32 & 0.46M & 400 & 57.5 & 62.0 & 69.8 \\
LiVT~\citep{xu2023livt} & ViT-B/16 & 85.80M & 100 & 58.2 & - & 69.2 \\
\hline
\multicolumn{7}{l}{\textbf{Fine-tuning pre-trained model}} \\
\hline
BALLAD~\citep{ma2021ballad} & ViT-B/16 & 149.62M & 50+10 & 77.8 & - & - \\
LIFT~\citep{shi2023lift} & ViT-B/16 & 0.10M & 10 & \underline{81.7} & \underline{83.1} & \underline{84.9} \\
\rowcolor{gray!20} Sage & ViT-B/16 & 6.96M & 10 & \textbf{83.3} & \textbf{84.6} & \textbf{86.3} \\
\hline
\end{tabular}
\label{tab:cifar}
\end{table*}
\subsection{Datasets}\label{sec:exp_dataset}
We perform long-tailed image classification experiments on four benchmark datasets: CIFAR-100-LT~\citep{cui2019class}, Places-LT~\citep{liu2019large}, ImageNet-LT~\citep{liu2019large}, and iNaturalist 2018~\citep{van2018inaturalist}.

\textbf{CIFAR-100-LT} is derived from CIFAR dataset~\citep{krizhevsky2009cifar}, containing 50,000 training and 10,000 validation images, with totally 100 categories. Following established protocols~\citep{cao2019ldam, li2022gcl}, we generate long-tailed distributions with imbalance ratios $\beta$ of 100, 50, and 10.

\textbf{ImageNet-LT} is the imbalanced variant of ImageNet dataset~\citep{russakovsky2015imagenet}. ImageNet-LT encompasses 115.8k images across 1,000 classes, with class sizes ranging from 1,280 to 5 samples, namely, $\beta=256$.

\textbf{Places-LT} is also the long-tailed version of the respective original dataset Places~\citep{zhou2017places}. It consists of 62,500 samples and exhibits a more pronounced imbalance with $\beta=996$, where class sizes range from 4,980 to 5.

\textbf{iNaturalist 2018} is the largest dataset in our evaluation. It contains 437.5k samples distributed across 8,142 categories. The class distribution ranges from 1,000 samples in the largest class to just 2 in the smallest, resulting in $\beta=500$.

\subsection{Implementation Details}\label{sec:exp_implement}
Consistent with previous works~\citep{zhong2021mislas, li2022gcl, du2024proco, shi2023lift}, we adopt top-1 classification accuracy as the evaluation metric to assess the performance of various methods. Additionally, we report the top-1 accuracy for head, medium, and tail classes to enable a more in-depth analysis. Following LIFT~\citep{shi2023lift}, we utilize the ViT-B/16 backbone for fine-tuning and employ LA loss~\citep{menon2020la} which is combined with our proposed compensation factor in Section~\ref{sec:loss}, as the loss function. For optimization, we use the SGD optimizer with a batch size of 128, a learning rate of 0.01, and a momentum of 0.90, alongside a cosine learning rate scheduler. The total number of training epochs is set to 10, except for the iNaturalist 2018 dataset, where it is set to 20. Our code is implemented with Pytorch and all experiments are carried out on an NVIDIA GeForce RTX 4080, except for the iNaturalist 2018 dataset which is conducted on an NVIDIA A100 80GB PCIe. All the hyperparameter settings used in our experiments are summarized in Appendix~\ref{ap:hyper_setting}.
\begin{table*}[t]
\centering
\setlength{\tabcolsep}{1.5pt}
\caption{Comparison on Places-LT in terms of top-1 accuracy (\%).}
\label{tab:places_lt}
\begin{tabular}{l|c|c|c|cccc}
\hline
\multirow{2}{*}{\textbf{Methods}} & \multirow{2}{*}{\textbf{Backbone}} & \textbf{Learnable} & \multirow{2}{*}{\textbf{Epochs}} & \multirow{2}{*}{\textbf{All}} & \multirow{2}{*}{\textbf{Head}} & \multirow{2}{*}{\textbf{Med.}} & \multirow{2}{*}{\textbf{Tail}} \\
& & \textbf{Params.} & & & & & \\
\hline
\multicolumn{8}{l}{\textbf{Training from scratch (with an ImageNet-1K pre-trained backbone)}} \\
\hline
OLTR~\citep{liu2019large} & ResNet-152 & 58.14M & 30 & 35.9 & 44.7 & 37.0 & 25.3 \\
cRT~\citep{kang2019decoupling} & ResNet-152 & 58.14M & 90+10 & 36.7 & 42.0 & 37.6 & 31.0 \\
LWS~\citep{kang2019decoupling} & ResNet-152 & 58.14M & 90+10 & 37.6 & 40.6 & 39.1 & 28.6 \\
MiSLAS~\citep{zhong2021mislas} & ResNet-152 & 58.14M & 90+10 & 40.4 & 39.6 & 43.3 & 36.1 \\
DisAlign~\citep{zhang2021disalign} & ResNet-152 & 58.14M & 30 & 39.3 & 40.4 & 42.4 & 34.0 \\
ALA~\citep{zhao2022adaptive} & ResNet-152 & 58.14M & 30 & 40.1 & 43.9 & 40.1 & 32.9 \\
PaCo~\citep{cui2021paco} & ResNet-152 & 58.14M & 30 & 41.2 & 36.1 & 47.9 & 35.3 \\
LiVT~\citep{xu2023livt} & ViT-B/16 & 85.80M & 100 & 40.8 & 48.1 & 40.6 & 27.5 \\
\hline
\multicolumn{8}{l}{\textbf{Fine-tuning foundation model}} \\
\hline
BALLAD~\citep{ma2021ballad} & ViT-B/16 & 149.62M & 50+10 & 49.5 & 49.3 & 50.2 & 48.4 \\
Decoder~\citep{wang2024decoder} & ViT-B/16 & 21.26M & $\sim$34 & 46.8 & - & - & - \\
LPT~\citep{dong2022lpt} & ViT-B/16 & 1.01M & 40+40 & 51.9 & \textbf{53.2} & 52.3 & 46.9 \\
LIFT~\citep{shi2023lift} & ViT-B/16 & 0.18M & 10 & \underline{52.2} & 51.7 & \textbf{53.1} & \underline{50.9} \\
\rowcolor{gray!20} Sage & ViT-B/16 & 7.13M & 10 & \textbf{52.7} & \underline{51.8} & \underline{52.8} & \textbf{53.8} \\
\hline
\end{tabular}
\end{table*}

\begin{table*}[t]
\centering
\setlength{\tabcolsep}{1.5pt}
\caption{Comparison on ImageNet-LT in terms of top-1 accuracy (\%). }
\label{tab:imagenet}
\begin{tabular}{l|c|c|c|cccc}
\hline
\multirow{2}{*}{\textbf{Methods}} & \multirow{2}{*}{\textbf{Backbone}} & \textbf{Learnable} & \multirow{2}{*}{\textbf{Epochs}} & \multirow{2}{*}{\textbf{All}} & \multirow{2}{*}{\textbf{Head}} & \multirow{2}{*}{\textbf{Med.}} & \multirow{2}{*}{\textbf{Tail}} \\
& & \textbf{Params.} & & & & & \\
\hline
\multicolumn{8}{l}{\textbf{Training from scratch}} \\
\hline
cRT~\citep{kang2019decoupling} & ResNet-50 & 23.51M & 90+10 & 47.3 & 58.8 & 44.0 & 26.1 \\
LWS~\citep{kang2019decoupling} & ResNet-50 & 23.51M & 90+10 & 47.7 & 57.1 & 45.2 & 29.3 \\
CDB-W-CE~\citep{sinha2022cdb} & ResNet-10 & 7.15M & 100 & 38.5 & - & - & -\\ 
MiSLAS~\citep{zhong2021mislas} & ResNet-50 & 23.51M & 180+10 & 52.7 & 62.9 & 50.7 & 34.3 \\
LA~\citep{menon2020la} & ResNet-50 & 23.51M & 90 & 51.1 & - & - & - \\
DisAlign~\citep{zhang2021disalign} & ResNet-50 & 23.51M & 90 & 52.9 & 61.3 & 52.2 & 31.4 \\
BCL~\citep{zhu2022bcl} & ResNet-50 & 23.51M & 100 & 56.0 & - & - & - \\
PaCo~\citep{cui2021paco} & ResNet-50 & 23.51M & 400 & 57.0 & - & - & - \\
NCL~\citep{li2022nested} & ResNet-50 & 23.51M & 400 & 57.4 & - & - & - \\
FUR~\citep{ma2024fur} & ResNext-50 & 25.03M & 100+50+50 & 55.5 & 65.4 & 52.2 & 37.8 \\
ProCo~\citep{du2024proco} & ResNet-50 & 23.51M & 180 & 57.8 & 68.2 & 55.1 & 38.1 \\
DBM-GML~\citep{son2024difficulty} & ResNet-50 & 23.51M & 90 & 57.4 & 65.3 & 55.1 & 43.1 \\
LiVT~\citep{xu2023livt} & ViT-B/16 & 85.80M & 100 & 60.9 & 73.6 & 56.4 & 41.0 \\
\hline
\multicolumn{8}{l}{\textbf{Fine-tuning foundation model}} \\
\hline
BALLAD~\citep{ma2021ballad} & ViT-B/16 & 149.62M & 50+10 & 75.7 & 79.1 & 74.5 & 69.8 \\
Decoder~\citep{wang2024decoder} & ViT-B/16 & 21.26M & $\sim$18 & 73.2 & - & - & - \\
LIFT~\citep{shi2023lift} & ViT-B/16 & 0.62M & 10 & \underline{78.3} & \underline{81.3} & \textbf{77.4} & \underline{73.4} \\
\rowcolor{gray!20} Sage & ViT-B/16 & 8.19M & 10 & \textbf{78.8} & \textbf{82.1} & \underline{77.3} & \textbf{74.8} \\
\hline
\end{tabular}
\end{table*}

\begin{table*}[t]
\centering
\setlength{\tabcolsep}{1.5pt}
\caption{Comparison on iNaturalist 2018 in terms of top-1 accuracy (\%).}
\label{tab:inat2018}
\begin{tabular}{l|c|c|c|cccc}
\hline
\multirow{2}{*}{\textbf{Methods}} & \multirow{2}{*}{\textbf{Backbone}} & \textbf{Learnable} & \multirow{2}{*}{\textbf{Epochs}} & \multirow{2}{*}{\textbf{All}} & \multirow{2}{*}{\textbf{Head}} & \multirow{2}{*}{\textbf{Med.}} & \multirow{2}{*}{\textbf{Tail}} \\
& & \textbf{Params.} & & & & & \\ 
\hline
\multicolumn{8}{l}{\textbf{Training from scratch}} \\
\hline
cRT~\citep{kang2019decoupling} & ResNet-50 & 23.51M & 90+10 & 65.2 & 69.0 & 66.0 & 63.2 \\
LWS~\citep{kang2019decoupling} & ResNet-50 & 23.51M & 90+10 & 65.9 & 65.0 & 66.3 & 65.5 \\
MiSLAS~\citep{zhong2021mislas} & ResNet-50 & 23.51M & 200+30 & 71.6 & 73.2 & 72.4 & 70.4 \\
DiVE~\citep{he2021dive} & ResNet-50 & 23.51M & 90 & 69.1 & 70.6 & 70.0 & 67.6 \\
DisAlign~\citep{zhang2021disalign} & ResNet-50 & 23.51M & 90 & 69.5 & 71.6 & 70.8 & 69.9 \\
ALA~\citep{zhao2022adaptive} & ResNet-50 & 23.51M & 90 & 70.7 & 71.3 & 70.8 & 70.4 \\
RIDE~\citep{wang2020long} & ResNet-50 & 23.51M & 100 & 72.6 & 70.9 & 72.4 & 73.1 \\
RIDE+CR~\citep{ma2023curvature} & ResNet-50 & 23.51M & 200 & 73.5 & 71.0 & 73.8 & 74.3 \\
RIDE+OTmix~\citep{gao2023enhancing} & ResNet-50 & 23.51M & 210 & 73.9 & 71.3 & 72.8 & 73.8 \\
BCL~\citep{zhu2022bcl} & ResNet-50 & 23.51M & 100 & 71.8 & - & - & - \\
PaCo~\citep{cui2021paco} & ResNet-50 & 23.51M & 100 & 73.2 & 70.4 & 72.8 & 73.6 \\
NCL~\citep{li2022nested} & ResNet-50 & 23.51M & 100 & 74.5 & 72.0 & 74.9 & 73.8 \\
GML~\citep{suh2023long} & ResNet-50 & 23.51M & 100 & 74.8 & - & - & - \\
FUR~\citep{ma2024fur} & ResNet-50 & 23.51M & {\small{100+50+50}} & 72.6 & 73.6 & 72.9 & 73.1 \\
ProCo~\citep{du2024proco} & ResNet-50 & 23.51M& 400 & 75.8 & \underline{74.0} & 76.0 & 76.0 \\
DBM-GML~\citep{son2024difficulty} & ResNet-50 & 23.51M & 100 & 72.0 & 66.9 & 71.9 & 73.6\\
LiVT~\citep{xu2023livt} & ViT-B/16 & 85.80M & 100 & 76.1 & \textbf{78.9} & 76.5 & 74.8 \\
\hline
\multicolumn{8}{l}{\textbf{Fine-tuning foundation model}} \\
\hline
Decoder~\citep{wang2024decoder} & ViT-B/16 & 21.26M & $\sim$5 & 59.2 & - & - & - \\
LPT~\citep{dong2022lpt} & ViT-B/16 & 1.01M & 80+80 & 76.1 & - & - & 79.3 \\
LIFT~\citep{shi2023lift} & ViT-B/16 & 4.75M & 20 & \underline{80.4} & \underline{74.0} & \underline{80.3} & \underline{82.2} \\
\rowcolor{gray!20} Sage & ViT-B/16 & 18.04M & 20 & \textbf{80.9} & \underline{74.0} & \textbf{80.6} & \textbf{83.1} \\
\hline
\end{tabular}
\end{table*}

\subsection{Comparison methods and results}\label{sec:exp_comparison}
We conduct comparative experiments with both deep neural network-based and foundation model-based long-tailed methods. The evaluated baselines and state-of-the-art approaches include:
(1) deep neural network-based methods: LDAM~\citep{cao2019ldam}, BBN~\citep{zhou2019bbn}, CDB-W-CE~\citep{sinha2022cdb}, DiVE~\citep{he2021dive}, cRT~\citep{kang2019decoupling}, MiSLAS~\citep{zhong2021mislas}, BS~\citep{ren2020bs}, PaCo~\citep{cui2021paco}, BCL~\citep{zhu2022bcl}, ALA~\citep{zhao2022adaptive}, RIDE~\citep{wang2020long}, CR~\citep{ma2023curvature}, OTmix~\citep{gao2023enhancing}, ProCo~\citep{du2024proco}, LWS~\citep{kang2019decoupling}, LA~\citep{menon2020la}, DisAlign~\citep{zhang2021disalign}, NCL~\citep{li2022nested}, GML~\citep{suh2023long}, FUR~\citep{ma2024fur}, DBM~\citep{son2024difficulty}, LiVT~\citep{xu2023livt},
(2) foundation model-based long-tailed methods:  BALLAD~\citep{ma2021ballad}, LPT~\citep{dong2022lpt}, Decoder~\citep{wang2024decoder}, LIFT~\citep{shi2023lift}.
We exclude the foundation model-based methods that leverage additional external data for a fair comparison.

Table~\ref{tab:cifar}, Table~\ref{tab:places_lt}, Table~\ref{tab:imagenet}, and Table~\ref{tab:inat2018} present the performances of competing approaches, with top-1 accuracy on the test sets serving as the evaluation metric. All baseline results are directly taken from~\citep{shi2023lift} and the respective publications, where `-' indicates that the corresponding results were not reported in the original paper. The best-performing method is \textbf{bolded}, while the second-best is \underline{underlined}.

\subsubsection{Comparison results on CIFAR-100-LT}
We perform experiments on CIFAR-100-LT with imbalance ratios of 100, 50, and 10, and the corresponding results are summarized in Table~\ref{tab:cifar}. By introducing a few learnable parameters to encode semantic information, our method delivers performance improvements of 1.6\%, 1.5\%, and 1.4\% on imbalance ratios of 100, 50, and 10, respectively, compared to the state-of-the-art method, LIFT. Compared with the majority of the ViT-B/16-based methods, such as LiVT and BALLAD, our approach occupied less computational resources and reduced training time, making it more efficient for practical applications. The necessity of the increased learnable parameters, along with their contribution to the overall performance, are comprehensively discussed in the Appendix~\ref{sec:ap_param}.

\subsubsection{Comparison on large-scale datasets}
The comparative results on the Places-LT, ImageNet-LT, and iNaturalist 2018 datasets are presented in Table~\ref{tab:places_lt}, Table~\ref{tab:imagenet}, and Table~\ref{tab:inat2018}, respectively. Across all large-scale datasets, our method improves overall performance, with significant enhancements for the tail class while keeping the performance for the head class. Specifically, as shown in Table~\ref{tab:places_lt} for the Places-LT dataset, our method boosts the performance of the tail class to 53.8\%, a 2.9\% improvement compared to the second-best method. For the ImageNet-LT dataset, detailed in Table~\ref{tab:imagenet}, our method achieves a 1.4\% improvement on the tail class, while simultaneously enhancing the performance on the head class by 0.8\%. These balanced improvements lead to a 0.5\% overall accuracy increase. Similarly, as indicated in Table~\ref{tab:inat2018} for the iNaturalist 2018 dataset, our method achieves a 0.9\% improvement for tail class while keeping the performance of head class. The analysis of the number of learnable parameters is provided in Appendix~\ref{sec:ap_param}.

\subsection{Ablation studies}\label{sec:exp_ablation}
In this section, we conduct a thorough analysis of Sage. In particular, we perform ablation studies on each component of Sage, as described in Section~\ref{sec:exp_abla_result}, to assess their respective contributions. The sensitivity of the introduced hyperparameters is analyzed in Section~\ref{sec:exp_sensitive}, and the effectiveness of the proposed compensation factor $\{\Lambda_i\}_{i=1}^C$ is further examined in Section~\ref{sec:exp_abla_cf_others}.

\begin{table*}[t]
\centering
\setlength{\tabcolsep}{2.5pt}
\caption{Ablation studies on CIFAR-100-LT dataset in terms of top-1 accuracy (\%).}
\begin{tabular}{cccc|cccc|cccc|cccc}
\hline
\multirow{2}{*}{\textbf{SG}} & \multirow{2}{*}{\textbf{Init}} & \multirow{2}{*}{\textbf{CF}} & \multirow{2}{*}{\textbf{FIT}} &
\multicolumn{4}{c|}{$\beta=100$} & 
\multicolumn{4}{c|}{$\beta=50$} & 
\multicolumn{4}{c}{$\beta=10$} \\ 
& & & & \textbf{All} & \textbf{Head} & \textbf{Med.} & \textbf{Tail} 
& \textbf{All} & \textbf{Head} & \textbf{Med.} & \textbf{Tail} 
& \textbf{All} & \textbf{Head} & \textbf{Med.} & \textbf{Tail}\\ 
\hline
 & & & & 81.7 & 85.2 & 82.4 & 76.7 & 83.1 & 85.3 & 82.6 & 81.4 & 84.9 & 86.2 & 83.7 & 84.6\\
\checkmark & & & & 81.9 & \underline{86.2} & 82.0 & 76.8 & \underline{84.1} & \underline{86.7} & 83.3 & 81.8 & 86.0 & \underline{87.2} & 84.9 & 85.9\\
\checkmark & \checkmark & & & 82.5 & \underline{86.2} & \textbf{83.1} & 77.2 & 83.9 & \textbf{86.8} & \underline{83.4} & 81.1 & 86.1 & \textbf{87.5} & \textbf{85.2} & 85.6\\
\checkmark & \checkmark & \checkmark & & \underline{82.6} & 85.8 & \underline{83.0} & \underline{78.3} & \underline{84.1} & 86.3 & 83.3 & \underline{82.5} & \underline{86.2} & \underline{87.2} & \underline{85.1} & \textbf{86.3} \\
\checkmark & \checkmark & \checkmark & \checkmark & \textbf{83.3} & \textbf{86.5} & 82.8 & \textbf{80.0} & \textbf{84.6} & \textbf{86.8} & \textbf{83.6} & \textbf{83.2} & \textbf{86.3} & \textbf{87.5} & \underline{85.1} & \textbf{86.3}\\
\hline
\end{tabular}
\label{tab:abla_cifar}
\end{table*}
\begin{table*}[t]
\centering
\setlength{\tabcolsep}{2.5pt}
\caption{Ablation studies across large-scale datasets in terms of top-1 accuracy (\%).}
\begin{tabular}{cccc|cccc|cccc|cccc}
\hline
\multirow{2}{*}{\textbf{SG}} & \multirow{2}{*}{\textbf{Init}} & \multirow{2}{*}{\textbf{CF}} & \multirow{2}{*}{\textbf{FIT}} &
\multicolumn{4}{c|}{\textbf{ImageNet-LT}} & 
\multicolumn{4}{c|}{\textbf{Places-LT}} & 
\multicolumn{4}{c}{\textbf{iNaturalist 2018}} \\ 
& & & & \textbf{All} & \textbf{Head} & \textbf{Med.} & \textbf{Tail} 
& \textbf{All} & \textbf{Head} & \textbf{Med.} & \textbf{Tail} 
& \textbf{All} & \textbf{Head} & \textbf{Med.} & \textbf{Tail}\\ 
\hline
 & & & & 78.3 & 81.3 & \underline{77.4} & 73.4 & 52.2 & 51.7 & \underline{53.1} & 50.9 & 80.4 & 74.0 & 80.3 & 82.2\\
\checkmark & & & & 78.4 & \underline{81.7} & 77.3 & 72.8 & 52.4 & \textbf{52.0} & 53.0 & 51.7 & \underline{80.8} & \textbf{75.4} & \underline{80.6} & 82.4\\
\checkmark & \checkmark & & & 78.4 & \underline{81.7} & \underline{77.4} & 72.9 & 52.5 & \textbf{52.0} & 53.0 & 52.2 & \textbf{80.9} & \underline{75.1} & \textbf{80.7} & 82.5\\
\checkmark & \checkmark & \checkmark & & \underline{78.5} & 81.2 & \textbf{77.5} & \underline{74.0} & \underline{52.6} & 51.3 & \textbf{53.3} & \underline{53.2} & \textbf{80.9} & 73.8 & \underline{80.6} & \underline{83.0}\\
\checkmark & \checkmark & \checkmark & \checkmark & \textbf{78.8} & \textbf{82.1} & 77.3 & \textbf{74.8} & \textbf{52.7} & \underline{51.8} & 52.8 & \textbf{53.8} & \textbf{80.9} & 74.0 & \underline{80.6} & \textbf{83.1}\\
\hline
\end{tabular}
\label{tab:abla_large}
\end{table*}
\subsubsection{Ablation experiment}\label{sec:exp_abla_result}
To gain a deeper understanding of the efficacy of each component in the proposed Sage, we conduct ablation experiments on individual components across all datasets. The components involved are summarized as follows: \\
(1) \textbf{SG} : refer to the proposed SG-Adapter within visual encoder, as detailed in Section~\ref{sec:sg-adapter}, \\
(2) \textbf{Init} : initialize using the average of the textual description from multiple templates, the involved templates derived are summarized in Appendix~\ref{ap:templates},\\
(3) \textbf{CF} : the ditribution mismatch-aware compensation factor illustrated in Section~\ref{sec:loss},\\
(4) \textbf{FIT} : refer to the feature interchange technique, as described in Section~\ref{sec:recall}. \\
The ablation results are presented in Table~\ref{tab:abla_cifar} and Table~\ref{tab:abla_large} for CIFAR-100-LT and large-scale datasets, respectively. The proposed SG-Adapter achieves significant improvements on CIFAR-100-LT across all imbalance factors. Specifically, it achieves overall improvements of 1.1\%, 1.0\%, and 0.2\% on CIFAR-100-LT with imbalance factors of 10, 50, and 100, respectively, by introducing only the SG-Adapter. Consistent enhancements are also observed on large-scale datasets, further highlighting the effectiveness of SG-Adapter.

Besides, the introduced compensation factor improves the performance of the tail class by 1.1\%, 1.4\%, and 0.7\% on CIFAR-100-LT with imbalance factors of 100, 50, and 10, respectively. And it achieves enhancements of 1.1\%, 1.0\%, and 0.5\% for the tail class on the ImageNet-LT, Places-LT, and iNaturalist 2018 datasets, respectively. These results emphasize the effectiveness of the compensation factor in mitigating biased predictions, thereby promoting performance improvement for the less frequent classes. Furthermore, by leveraging FIT, the performance of the tail class can be further improved. Specifically, FIT achieves gains of 1.7\% and 0.7\% on the CIFAR-100-LT dataset with imbalance factors of 100 and 50, respectively, with enhanced performance on the head class. On large-scale datasets, FIT further enhances the performance of the tail class up to 0.8\%, while greatly maintaining the performance of the head class. Specifically, FIT promotes a further 0.9\% enhancement on the head class on ImageNet-LT. In summary, all the results demonstrate that each component effectively enhances the performance.
\begin{figure*}[!t]
\centering
\includegraphics[scale=0.26]{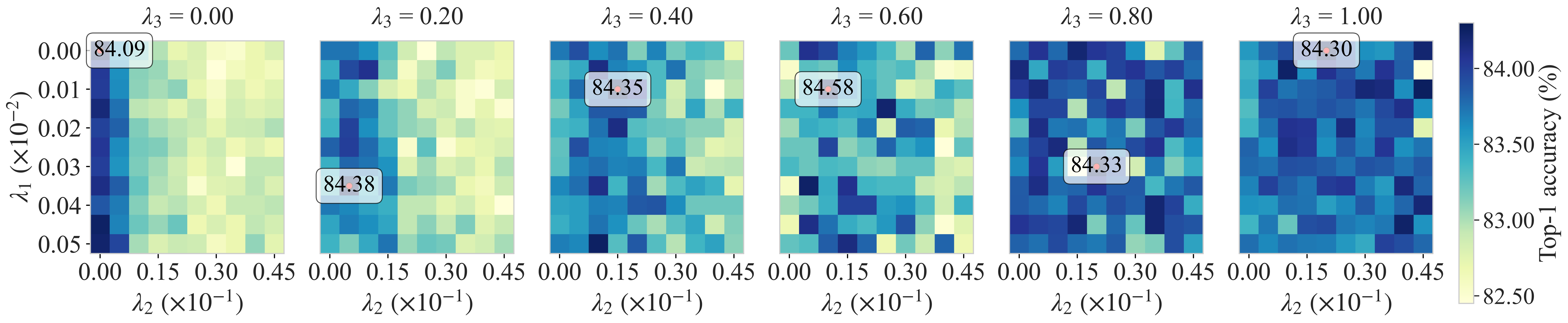}
\caption{Parameter analysis of $\lambda_1$, $\lambda_2$, and $\lambda_3$ utilized in the FIT module conducted on the CIFAR-100-LT dataset with $\beta=50$.}
\label{fig:man_init_zs}
\end{figure*}

\begin{figure}[!t]
\centering
\includegraphics[scale=0.385]{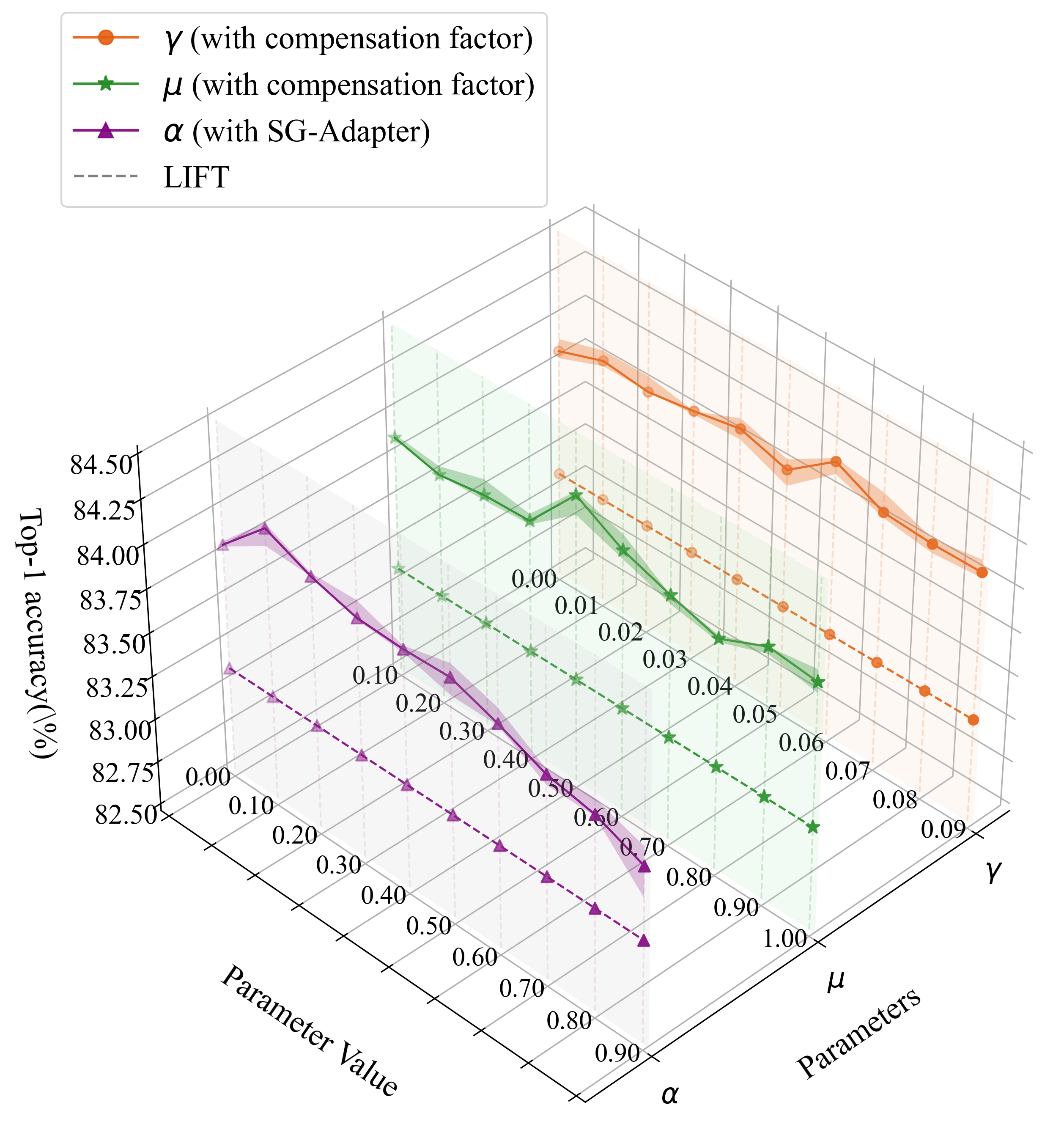}
\caption{Parameter analysis of $\gamma$, $\mu$, and $\alpha$ on CIFAR-100-LT dataset with $\beta=50$, without FIT as stated in Section~\ref{sec:recall}.}
\label{fig:gamma_mu_alpha}
\end{figure}

\subsubsection{Sensitive studies}\label{sec:exp_sensitive}
In this section, we delve into the sensitivity of introduced hyperparameters, including $\alpha$ in SG-Adapter (please refer to Section~\ref{sec:sg-adapter}),  $\mu$ and $\gamma$ in the proposed compensation factor $\{\Lambda_i\}_{i=1}^C$ (please refer to Section~\ref{sec:loss}), and hyperparameters $\lambda_1$, $\lambda_2$, and $\lambda_3$ control the influence of multiple loss functions in FIT (please refer to Section~\ref{sec:recall}).

The impact of the hyperparameters $\mu$ and $\gamma$ on the compensation factor $\{\Lambda_i\}_{i=1}^C$, as well as the influence of $\alpha$ on the SG-Adapter, is summarized in Fig.~\ref{fig:gamma_mu_alpha}. Specifically, we emphasize the effect of $\gamma\in [0.00, 0.09]$ and $\mu \in [0.10, 1.00]$. Since $\mu$ and $\gamma$ serve as approximations of $\frac{P^{\prime}_s(\mathbf{x}_i)}{P_t(\mathbf{x}_i)} \approx \mu n_i^{\gamma}$ for sample $\mathbf{x}$ from the $i$-th class, it is observed that the optimal performance is achieved when $\gamma = 0.06$ and $\mu = 0.50$. Moreover, the proposed method consistently outperforms the baseline across all settings, where the values providing the most precise approximation deliver the best performance.

For $\alpha$, which governs the contribution of negative activations, we analyze the produced influence when $\alpha \in [0.00, 0.90]$. The best result is achieved at $\alpha=0.10$, while performance decreases with $\alpha$ when $\alpha>0.10$. The results indicate that properly incorporating negative activations can improve performance, while excessive use of negative activations causes a reduction in performance.

\begin{table*}[t]
\centering
\setlength{\tabcolsep}{2.5pt}
\caption{Ablation studies of compensation factor combined with other models on the CIFAR-100-LT dataset in terms of top-1 accuracy (\%).}
\begin{tabular}{l|cccc|cccc|cccc}
\hline
\multirow{2}{*}{\textbf{Methods}} &
\multicolumn{4}{c|}{$\beta=100$} & 
\multicolumn{4}{c|}{$\beta=50$} & 
\multicolumn{4}{c}{$\beta=10$} \\ 
& \textbf{All} & \textbf{Head} & \textbf{Med.} & \textbf{Tail} 
& \textbf{All} & \textbf{Head} & \textbf{Med.} & \textbf{Tail} 
& \textbf{All} & \textbf{Head} & \textbf{Med.} & \textbf{Tail}\\ 
\hline
CE & 42.4 & \textbf{69.2} & 42.2 & 11.5 & 45.3 & \textbf{70.7} & \textbf{46.2} & 14.6 & 57.9 & \textbf{65.8} & 39.5 & 0.0 \\
\rowcolor{gray!20} CE + CF & \textbf{45.5} & 63.2 & \textbf{45.6} & \textbf{24.7} & \textbf{49.8} & 63.0 & 45.0 & \textbf{30.4} & \textbf{60.1} & 63.7 & \textbf{51.8} & 0.0 \\
\hline
KPS & 44.9 & 57.1 & 43.1 & 32.9 & 49.5 & 56.7 & 44.8 & 44.5 & 60.3 & 62.3 & 55.6 & 0.0 \\
\rowcolor{gray!20} KPS + CF & \textbf{45.8} & \textbf{57.2} & \textbf{44.8} & \textbf{33.5} & \textbf{50.6} & \textbf{57.7} & \textbf{45.6} & \textbf{45.4} & \textbf{60.9} & \textbf{62.4} & \textbf{57.4} & 0.0 \\
\hline
ProCo & 51.7 & \textbf{66.8} & 52.8 & 32.7 & 56.2 & \textbf{66.4} & 51.2 & 44.5 & 65.0 & \textbf{67.8} & 58.8 & 0.0\\
\rowcolor{gray!20} ProCo + CF & \textbf{52.3} & 66.1 & \textbf{53.6} & \textbf{34.6} & \textbf{56.7} & 66.0 & \textbf{52.5} & \textbf{45.2}  & \textbf{65.5} & 67.5 & \textbf{61.1} & 0.0 \\
\hline
\end{tabular}
\label{tab:abla_cf_others}
\end{table*}
As defined in Eq.~(\ref{eq:loss_final_final}), $\lambda_1$, $\lambda_2$, and $\lambda_3$ regulate the contributions of the loss function to different logits. Specifically, we analyze the case where $\lambda_1\in [0.0000, 0.0005]$, $\lambda_2\in [0.000, 0.045]$, and $\lambda_3\in [0.0, 1.0]$, as illustrated in Fig.~\ref{fig:man_init_zs}, with a particular emphasis on the optimal results for different values of $\lambda_3$. In most cases, the model achieves optimal performance when $\lambda_2 \leq 0.02$, with performance declining as $\lambda_2$ increases further, on the CIFAR-100-LT with $\beta=50$. Among the optimal values across different $\lambda_3$, the global optimum is reached at $\lambda_3=0.60$, beyond which performance also deteriorates. This suggests that moderately constraining the model to balance the generality of the foundation model and the task-specificity of the fine-tuned model is beneficial. However, overly emphasizing generality at the expense of task-specificity leads to performance degradation.

\subsubsection{Analysis on compensation factor}\label{sec:exp_abla_cf_others}
To evaluate the effectiveness and compatibility of the proposed compensation factor $\{\Lambda_i\}_{i=1}^C$ for all $C$ classes introduced in Section~\ref{sec:loss}, we combine compensation factor with various loss functions and models. The results in Table~\ref{tab:abla_cf_others} demonstrate that incorporating the compensation factor consistently improves overall performance across different loss functions and models, with a particularly notable impact on the medium and tail classes. For models trained using the CE loss, the inclusion of compensation factor results in an improvement of over 10.0\% in the tail class, reaching up to 15.8\% in some cases. When combined with compensation factor, KPS~\citep{li2022kps} achieves overall performance enhancements, including improvements of up to 1.8\% in the medium and tail classes. Moreover, integrating compensation factor provides ProCo yields performance gains ranging from 0.7\% to 2.3\% in the medium and tail classes.

\subsection{Limitations}\label{sec:exp_limitations}
Although SG-Adapter effectively incorprate the semantics into the process of fine-tuning visual encoder, the cost is the increased number of parameters brought by the projection matrices introduced with semantic features. The parameter quantity of each component in SG-Adapter is listed in Table~\ref{tab:ap_param}, from which we find that the multi-modal projection matrix $\mathbf{W}^{vt}_{\text{proj}}$ accounts for the largest portion of the parameter count. However, the exclusion of $\mathbf{W}^{vt}_{\text{proj}}$ limits the potential benefits for the tail class, based on our analysis in Appendix~\ref{sec:ap_param} and demonstrated in Table~\ref{tab:wo_wup_proj_cifar} and Table~\ref{tab:wo_wup_proj_large}. The future enhancement can focus on strengthening the alignment between visual and textual modalities while minimizing the introduction of extra learnable parameters.

On the other hand, similar to many deep learning methods, this work introduces several hyperparameters. The selection of optimal hyperparameters varies across different datasets. However, as demonstrated by the results in Section~\ref{sec:exp_sensitive}, our method consistently achieves top performance across nearly all configurations, highlighting the robustness of the proposed approach. Reducing the number of hyperparameters remains a potential avenue for future enhancement.

\section{Conclusion}
In this work, we have presented \textit{Sage}, a novel learning strategy designed to address the challenge of diminished alignment between the visual and textual modalities in advanced methods. Specifically, we have introduced SG-Adapter, which integrated semantic guidance derived from the textual modality during the fine-tuning of the visual encoder. The integrated guidance passed semantics into the attention mechanism within the visual encoder and significantly enhanced the focus of the model on semantically relevant content. Nevertheless, we observed that the inconsistent class-conditional distributions between the training and test data, which were overlooked by the existing loss function, resulted in the performance improvements of the tail class being fewer than those of the head class. Based on our theoretical analysis, the existing loss function has been improved by combining it with a distribution mismatch-aware compensation factor, which reduces the prediction bias caused by the neglected inconsistent class-conditional distributions and further promotes the performance improvement of the tail class. Extensive experiments on various benchmark datasets have validated the effectiveness of Sage, demonstrating its ability to achieve superior performance in long-tailed visual recognition.

\backmatter

\bmhead{Data Availability}
This work uses publicly available data. CIFAR-100 is available at \url{https://www.cs.toronto.edu/~kriz/cifar.html}; ImageNet-LT and Places-LT are available at \url{https://liuziwei7.github.io/projects/LongTail.html}; iNaturalist 2018 is available at \url{https://github.com/visipedia/inat_comp/tree/master/2018}. All source codes will be publicly available upon the acceptance of this paper.

\bibliography{refer}

\newpage


\begin{oldappendices}
    
\section{Related work}
\textbf{Long-tailed learning with deep neural networks.}
Advanced long-tailed methods with deep neural networks include logit adjustment techniques and two-stage-based methods. Logit adjustment aims to shift the margin of each class in order to leave a larger margin for the overwhelmed tail classes. LA~\citep{menon2020la} allocates the adjustment of margin proportional to the label frequency regarding the balanced error. KPS~\citep{li2022kps} assigns a large margin to the key points, which represent the most challenging instances to distinguish between categories and play a critical role in improving overall classification performance. DDC~\citep{wang2023ddc} conducts an in-depth analysis of the impact of hyperparameters on the unified loss function using local Lipschitz continuity and provides recommended settings for these hyperparameters. However, designing an optimal and efficient approach to determine the margins used in logit adjustment remains a significant challenge.

Additionally, the two-stage framework is one of the most popular architectures in long-tailed learning. This architecture typically involves two stages: In the first stage, the model is trained on the original long-tailed dataset, whereas in the second stage, the classifier is retrained on the resampled imbalanced data while keeping the feature extractor frozen. MiSLAS~\citep{zhong2021mislas} highlights that Mixup improves representation learning and proposes a label smoothing strategy to address imbalanced learning. GCL~\citep{li2022gcl} introduces Gaussian-shaped noise to simulate latent samples for the less frequent classes and applies an effective number-based re-sampling technique during the later stage. H2T~\citep{li2024h2t} combines data from both a balanced and an imbalanced branch in the second stage, effectively enriching the semantics of tail samples with the support of head samples. By calculating the class-wise distribution, FUR~\citep{ma2024fur} augments the less frequent classes based on the feature uncertainty representation, which is then used to re-train the classifier in the second stage. These approaches leverage the advantages of long-tailed data to train a robust feature extractor, while utilizing re-balanced data to optimize the classifier. However, these methods take longer to train and face challenges due to distribution differences in the training data across two stages.

\textbf{Long-tailed learning via foundation models.}
Due to its large amount of learned knowledge, the foundation model provides an efficient and useful way to rapidly satisfy the learning needs for certain downstream tasks. Nowadays, the pre-trained visual-language models have started to benefit the image classification task under long-tailed scenarios. BALLAD~\citep{ma2021ballad} first continues the pretraining of the vision-language model with long-tailed data and subsequently introduces a linear adapter to strengthen the representation learning of tail class with re-balanced data. TACKLE~\citep{ma2022TACKLE} leverages the strength of vision-language models by retrieving relevant images from the web, in order to address the imbalanced learning caused by insufficient class samples. ECVL~\citep{song2023ECVL} reframes visual recognition as a visual-language matching task by measuring the similarity between visual and text inputs within a two-stage framework. The first stage involves training with image features and class labels encoded by the respective encoders, while the second stage incorporates a balanced linear adapter to enable more targeted and refined training.

In contrast to training the whole model, fine-tuning techniques seem more efficient and avoid forgetting amounts of common knowledge stored in foundation models. VL-LTR~\citep{tian2022vlltr} pretrains the visual and text encoders using class-specific contrastive learning and utilizes high-scored sentences as anchors for classification. Instead of training the entire model, LPT~\citep{dong2022lpt} focuses on training shared and group-specific prompts. The shared prompt captures common knowledge across all classes, while the group-specific prompts learn distinguishing features from similar samples. Focusing on improving the performance of tail class in long-tailed learning through visual prompt tuning, GNM-PT~\citep{li2024improving} explores gradient descent directions within a random parameter neighborhood, independent of input samples during tuning, to enhance generalization across all classes. LIFT~\citep{shi2023lift} applies the advanced AdaptFormer~\citep{chen2022adaptformer} for long-tailed learning through adaptive lightweight fine-tuning, achieving excellent performance across all classes with low complexity and reduced training time. However, when adapting CLIP model~\citep{radford2021clip} to the downstream task, these methods fine-tune the visual encoder using unimodal visual features while neglecting the incorporation of semantics derived from the textual modality, weakening the alignment between the visual and textual modalities, therefore leading the model to focus incorrectly on irrelevant visual regions.

\section{Derivation in multi-head attention mechanism}\label{ap:multi_head}
Following the discussion in Section~\ref{sec:sg-adapter}, where $\dot{\mathbf{f}}=[\mathbf{f}_1, \mathbf{f}_2]$ and $\mathbf{W}_{q}^j$ as $\begin{bmatrix} \dot{\mathbf{W}}_{q}^j \\ \ddot{\mathbf{W}}_{q}^j \end{bmatrix}$. $\mathbf{Q}^j$, $\mathbf{K}^j$, and $\mathbf{V}^j$ in $\operatorname{head}_j$ can be presented as:
\begin{equation}
\begin{split}
\mathbf{Q}^j &=[\mathbf{f}_1\dot{\mathbf{W}}_{q}^j+\mathbf{f}_2\ddot{\mathbf{W}}_{q}^j],\\
\mathbf{K}^j&=[\mathbf{f}_1\dot{\mathbf{W}}_{k}^j+\mathbf{f}_2\ddot{\mathbf{W}}_{k}^j],\\
\mathbf{V}^j&=[\mathbf{f}_1\dot{\mathbf{W}}_{v}^j+\mathbf{f}_2\ddot{\mathbf{W}}_{v}^j].
\end{split}
\end{equation}
Therefore, excluding $\operatorname{Softmax}$ and the constant $\sqrt{d_k}$, Eq.~(\ref{eq:ori_multihead}) for $\operatorname{head}_j$ simplifies to:
\begin{equation}
\begin{split}
&\left(\mathbf{Q}^j {\mathbf{K}^j}^{\top}\right) \mathbf{V}^j\\
&=\left(\left(\mathbf{f}_1\dot{\mathbf{W}}_{q}^j+\mathbf{f}_2\ddot{\mathbf{W}}_{q}^j\right)\left(\mathbf{f}_1\dot{\mathbf{W}}_{k}^j+\mathbf{f}_2\ddot{\mathbf{W}}_{k}^j\right)\phantom{}^{\top}\right)\\
&\hspace{0.5cm} \cdot \left(\mathbf{f}_1\dot{\mathbf{W}}_{v}^j+\mathbf{f}_2\ddot{\mathbf{W}}_{v}^j\right)\\
&=\left( \mathbf{f}_1\dot{\mathbf{W}}_{q}^j \dot{\mathbf{W}}_{k}^j\phantom{}^{\top} {\mathbf{f}_1}\phantom{}^{\top} 
+ \mathbf{f}_1\dot{\mathbf{W}}_{q}^j \ddot{\mathbf{W}}_{k}^j\phantom{}^{\top} {\mathbf{f}_2}\phantom{}^{\top} 
+ \mathbf{f}_2\ddot{\mathbf{W}}_{q}^j \dot{\mathbf{W}}_{k}^j\phantom{}^{\top} {\mathbf{f}_1}\phantom{}^{\top} \right. \\
&\hspace{0.5cm} + \left. \mathbf{f}_2\ddot{\mathbf{W}}_{q}^j \ddot{\mathbf{W}}_{k}^j\phantom{}^{\top} {\mathbf{f}_2}\phantom{}^{\top} \right) \cdot
\left( \mathbf{f}_1\dot{\mathbf{W}}_{v}^j+\mathbf{f}_2{\ddot{\mathbf{W}}_{v}^j} \right)\\
&=\mathbf{f}_1\dot{\mathbf{W}}_{q}^j \dot{\mathbf{W}}_{k}^j\phantom{}^{\top} {\mathbf{f}_1}\phantom{}^{\top}{\mathbf{f}_1}\dot{\mathbf{W}}_{v}^j 
+ \mathbf{f}_1\dot{\mathbf{W}}_{q}^j \ddot{\mathbf{W}}_{k}^j\phantom{}^{\top} {\mathbf{f}_2}\phantom{}^{\top} {\mathbf{f}_1}\dot{\mathbf{W}}_{v}^j 
\\
&\hspace{0.5cm} + \mathbf{f}_2\ddot{\mathbf{W}}_{q}^j \dot{\mathbf{W}}_{k}^j\phantom{}^{\top} {\mathbf{f}_1}\phantom{}^{\top}{\mathbf{f}_1}\dot{\mathbf{W}}_{v}^j 
+ \mathbf{f}_2\ddot{\mathbf{W}}_{q}^j \ddot{\mathbf{W}}_{k}^j\phantom{}^{\top} {\mathbf{f}_2}\phantom{}^{\top}{\mathbf{f}_1}\dot{\mathbf{W}}_{v}^j \\
&\hspace{0.5cm} + \mathbf{f}_1\dot{\mathbf{W}}_{q}^j \dot{\mathbf{W}}_{k}^j\phantom{}^{\top} {\mathbf{f}_1}\phantom{}^{\top}{\mathbf{f}_2}\ddot{\mathbf{W}}_{v}^j
+ \mathbf{f}_1\dot{\mathbf{W}}_{q}^j \ddot{\mathbf{W}}_{k}^j\phantom{}^{\top} {\mathbf{f}_2}\phantom{}^{\top} {\mathbf{f}_2}\ddot{\mathbf{W}}_{v}^j
\\
&\hspace{0.5cm} + \mathbf{f}_2 \ddot{\mathbf{W}}_{q}^j \dot{\mathbf{W}}_{k}^j\phantom{}^{\top} {\mathbf{f}_1}\phantom{}^{\top}{\mathbf{f}_2} \ddot{\mathbf{W}}_{v}^j 
+ \mathbf{f}_2 \ddot{\mathbf{W}}_{q}^j \ddot{\mathbf{W}}_{k}^j\phantom{}^{\top} {\mathbf{f}_2}\phantom{}^{\top}{\mathbf{f}_2}\ddot{\mathbf{W}}_{v}^j.
\end{split}
\end{equation}
This concludes the derivation.

\section{Empirical study of approximation}\label{ap:psx_ptx_ratio}
In this section, we examine the rationality of approximating $\frac{P^{\prime}_s(\mathbf{x})}{P_t(\mathbf{x})}$ as $ \frac{P^{\prime}_s(\mathbf{x}_i)}{P_t(\mathbf{x}_i)} \approx \mu n_i^{\gamma}$, as discussed in Section~\ref{sec:loss}, where $\mathbf{x}_i$ denotes a specific sample $\mathbf{x}$ from the $i$-th class. Assuming that $\mathbf{x}\in\mathcal{X}$ follows a Gaussian distribution, the class-conditional distribution $P(\mathbf{x}\mid y)$ of $\mathbf{x}$ is expressed as:
\begin{equation}
\label{eq:ap_psx_ptx_1}
\begin{split}
P(\mathbf{x}\mid y) = \frac{1}{\sqrt{(2\pi)^d |\mathbf{\Sigma}_y|}} e^{-\frac{1}{2} (\mathbf{x} - \mathbf{m}_y)^{\top} \mathbf{\Sigma}_y^{-1} (\mathbf{x} - \mathbf{m}_y)},
\end{split}
\end{equation}
where $\mathbf{m}_y$ and $\mathbf{\Sigma}_y$ is the mean vector and variance of the $y$-th class, respectively. Then the marginal probability distribution $P(\mathbf{x})$ can be calculated as:
\begin{equation}
\label{eq:ap_psx_ptx_2}
\begin{split}
P(\mathbf{x}) = \sum_k P(k) P(\mathbf{x}\mid y=k),
\end{split}
\end{equation}
where $k\in\{1,2, \ldots, C\}$. Starting from the case of $C=2$ with sample $\mathbf{x}$ belonging to class $1$, Eq.~(\ref{eq:ap_psx_ptx_2}) can be reformulated as:
\begin{equation}
\label{eq:ap_psx_ptx_3}
\begin{split}
P(\mathbf{x}) &= P(1) P(\mathbf{x}\mid 1)+ P(2) P(\mathbf{x}\mid 2)\\
&= \frac{P(1)}{\sqrt{(2\pi)^d |\mathbf{\Sigma}_1|}}\cdot \frac{1}{e^{\frac{1}{2} (\mathbf{x} - \mathbf{m}_1)^{\top} \mathbf{\Sigma}_1^{-1} (\mathbf{x} - \mathbf{m}_1)}} \\
&+ \frac{P(2)}{\sqrt{(2\pi)^d |\mathbf{\Sigma}_2|}} \cdot \frac{1}{e^{\frac{1}{2} (\mathbf{x} - \mathbf{m}_2)^{\top} \mathbf{\Sigma}_2^{-1} (\mathbf{x} - \mathbf{m}_2)}}.
\end{split}
\end{equation}
\begin{figure*}[t!]
    \centering
    \subfigure[CIFAR-100-LT, $\beta=100$]{
        \includegraphics[scale=0.255]{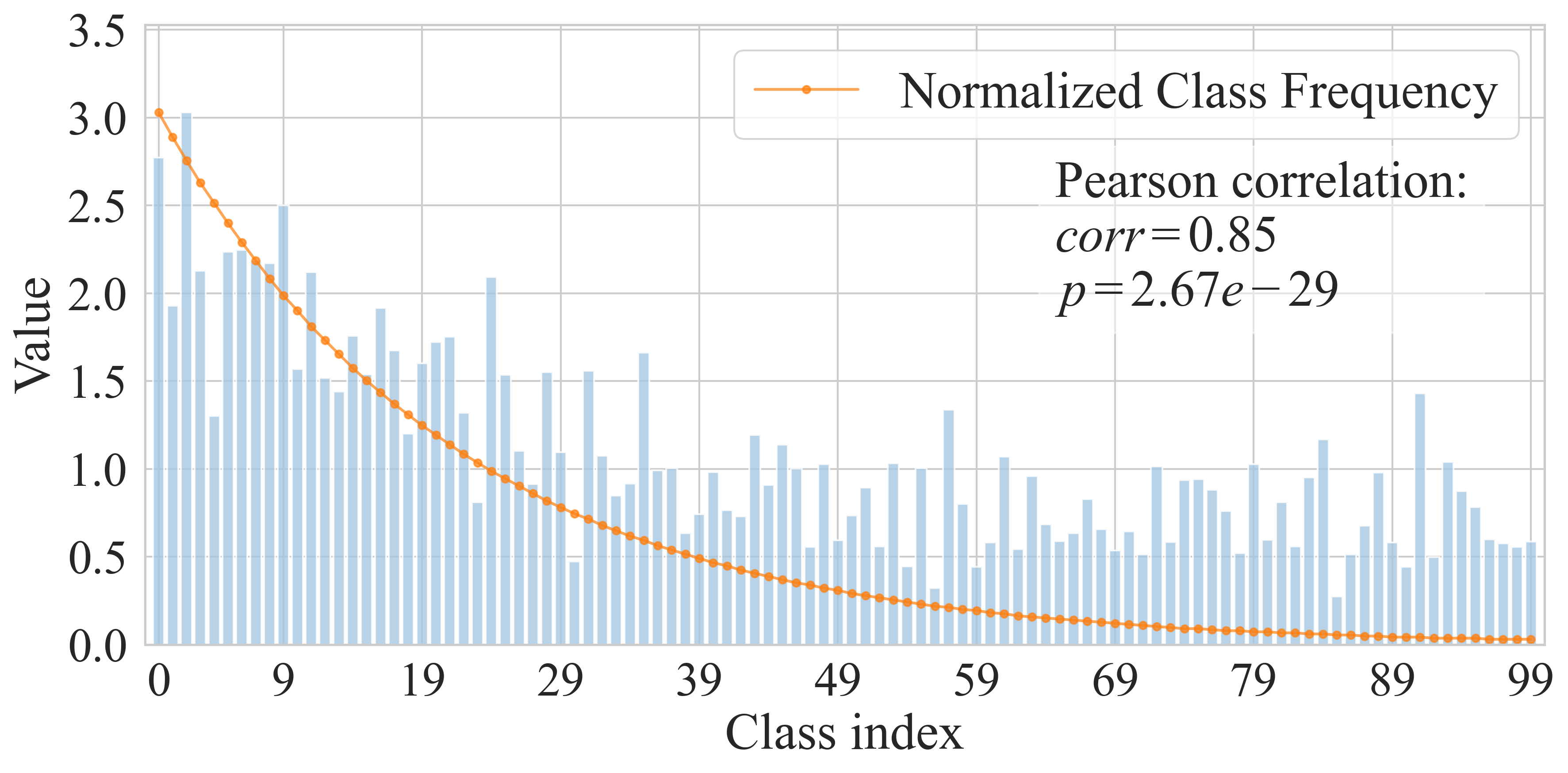} 
    }
    \subfigure[CIFAR-100-LT, $\beta=10$]{
        \includegraphics[scale=0.255]{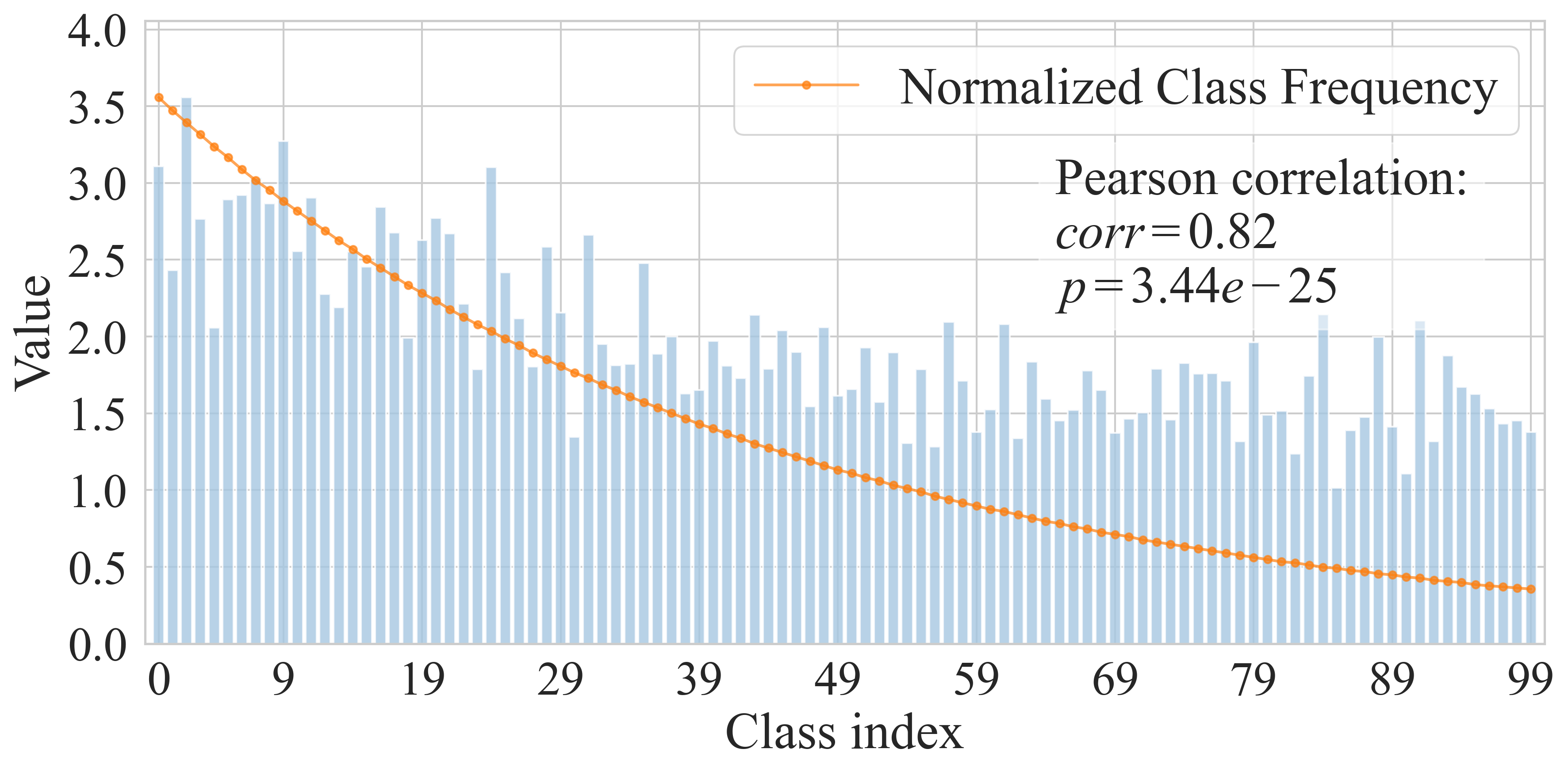} 
    }
    \subfigure[Places-LT]{
        \includegraphics[scale=0.255]{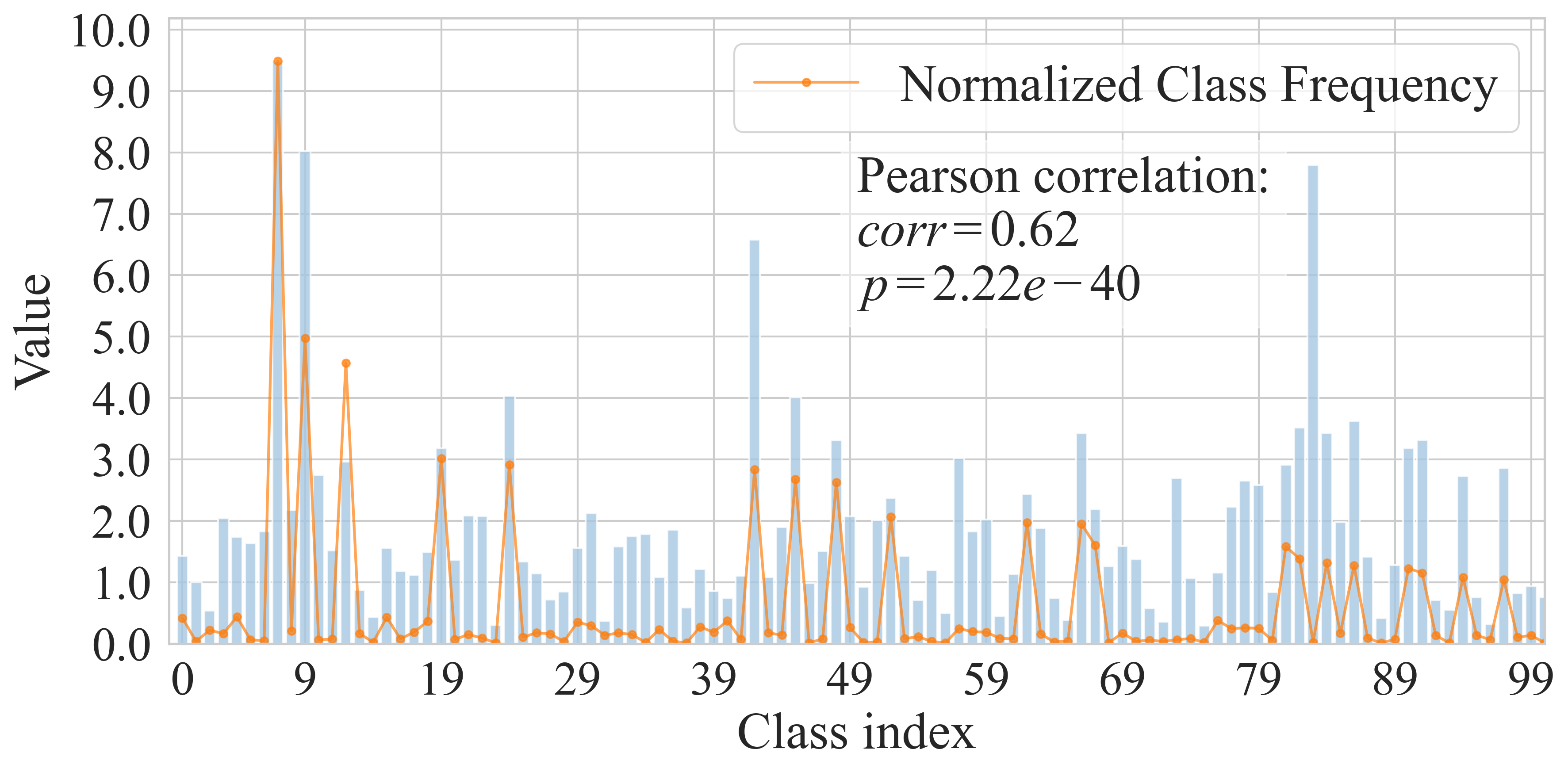} 
    }
    \subfigure[ImageNet-LT]{
        \includegraphics[scale=0.255]{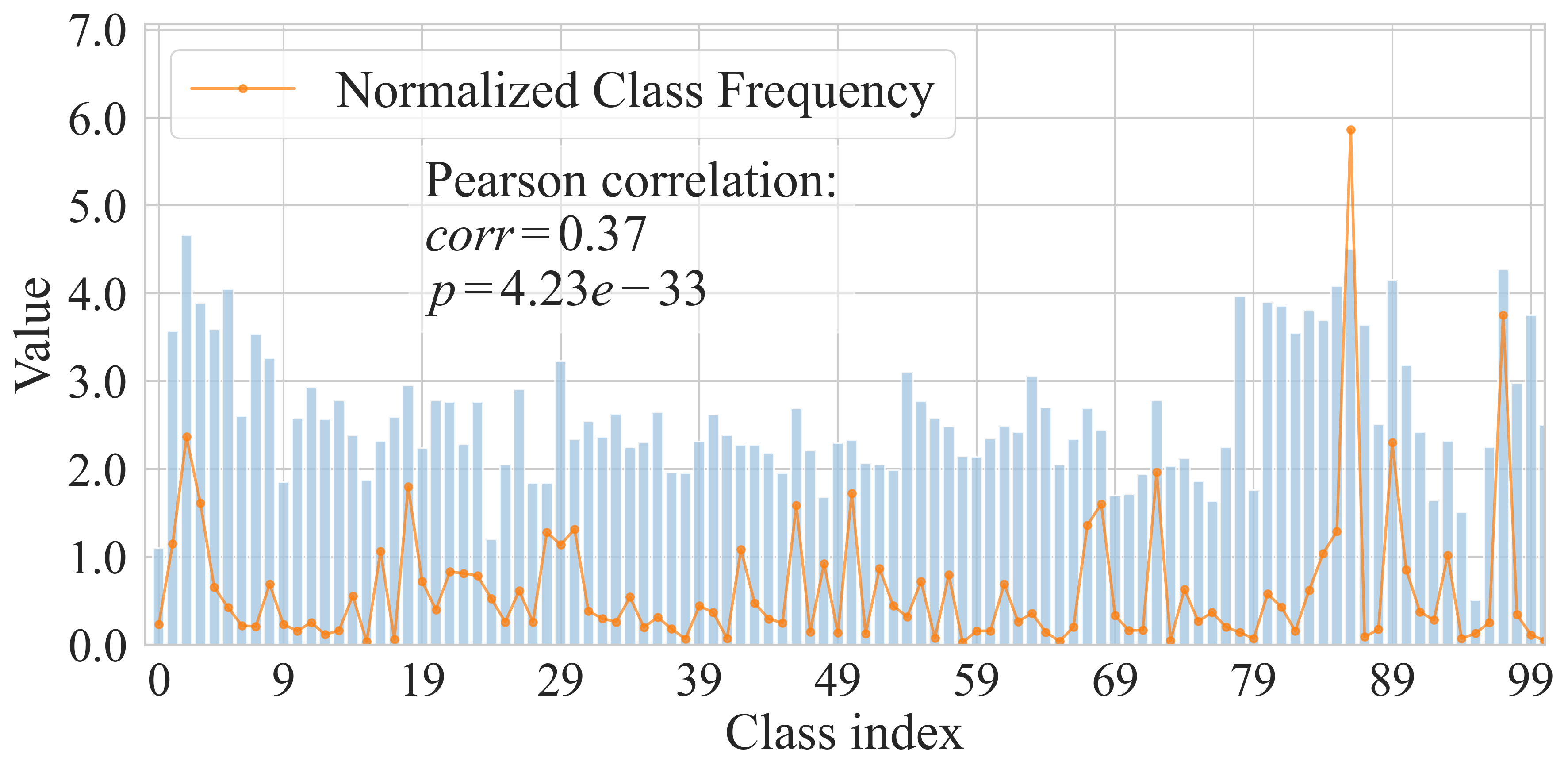}
    }
    \caption{Pearson correlation coefficient (denoted as `\textit{corr}') and p-value (denoted as `\textit{p}') between $\{\frac{P^{\prime}_s(\mathbf{x}_i)}{P_t(\mathbf{x}_i)}\}_{i=1}^C$ and the linear mapped size of each class across the datasets.}
    \label{fig:approximation}
\end{figure*}
Then, based on Eq.~(\ref{eq:ap_psx_ptx_3}), we have:
\begin{equation}
\label{eq:ap_psx_ptx_4}
\begin{split}
\frac{P(\mathbf{x})}{P(1)} 
&= \frac{1}{\sqrt{(2\pi)^d |\mathbf{\Sigma}_1|}}\cdot \frac{1}{e^{\frac{1}{2} (\mathbf{x} - \mathbf{m}_1)^{\top} \mathbf{\Sigma}_1^{-1} (\mathbf{x} - \mathbf{m}_1)}} \\
&+ \frac{P(2)}{P(1)}\cdot\frac{1}{\sqrt{(2\pi)^d |\mathbf{\Sigma}_2|}} \cdot \frac{1}{e^{\frac{1}{2} (\mathbf{x} - \mathbf{m}_2)^{\top} \mathbf{\Sigma}_2^{-1} (\mathbf{x} - \mathbf{m}_2)}}.
\end{split}
\end{equation}
The ideal classification model aims to minimize intra-class compactness and maximize inter-class dispersion~\citep{wen2016discriminative}, leading to
\begin{equation}
\label{eq:ap_psx_ptx_5}
\begin{split}
\frac{1}{\sqrt{(2\pi)^d |\mathbf{\Sigma}_1|}}\cdot \frac{1}{e^{\frac{1}{2} (\mathbf{x} - \mathbf{m}_1)^{\top} \mathbf{\Sigma}_1^{-1} (\mathbf{x} - \mathbf{m}_1)}}\rightarrow\tau_1, \\
\frac{1}{\sqrt{(2\pi)^d |\mathbf{\Sigma}_2|}} \cdot \frac{1}{e^{\frac{1}{2} (\mathbf{x} - \mathbf{m}_2)^{\top} \mathbf{\Sigma}_2^{-1} (\mathbf{x} - \mathbf{m}_2)}}\rightarrow\tau_2
\end{split}
\end{equation}
for the samples within the same class, where $\tau_1$ and $\tau_2$ are constants. Therefore, Eq.~(\ref{eq:ap_psx_ptx_4}) is transformed into the following form:
\begin{equation}
\label{eq:ap_psx_ptx_6}
\begin{split}
\frac{P(\mathbf{x})}{P(1)} 
= \tau_1 + \frac{P(2)}{P(1)}\cdot\tau_2 \quad \Rightarrow \quad \\
P(\mathbf{x})
= \left(\tau_1 + \frac{P(2)}{P(1)}\cdot\tau_2 \right) \cdot P(1).
\end{split}
\end{equation}

\begin{table*}[t]
  \centering
  \caption{Detailed experiment settings on benchmark datasets.}
  \begin{tabular}{l|ccc|ccc}
    \toprule
    Dataset & $\alpha$ & $\mu$ & $\gamma$ & $\lambda_1$ & $\lambda_2$ & $\lambda_3$ \\
    \midrule
    CIFAR-100-LT \hspace{0.3cm} $\beta$=100 & 0.100 & 0.500 & 0.050 & 0.0150000 & 0.0150000 & 0.4000000\\
    CIFAR-100-LT \hspace{0.3cm} $\beta$=50 & 0.100 & 0.500 & 0.060 & 0.0001000 & 0.0100000 & 0.6000000\\
    CIFAR-100-LT \hspace{0.3cm} $\beta$=10 & 0.100 & 0.500 & 0.060 & 0.0020000 & 0.0050000 & 0.2000000 \\
    \midrule
    Places-LT & 0.800 & 0.500 & 0.050 & 0.0030000 & 0.0100000 & 0.2000000\\
    ImageNet-LT & 0.800 & 0.500 & 0.050 & 0.0050000 & 0.0100000 & 0.2000000 \\
    iNaturalist 2018 & 1.000 & 0.500 & 0.085 & 0.0000001 & 0.0000001 & 0.0000001\\
    \bottomrule
  \end{tabular}
\label{tab:setting}
\end{table*}

\begin{table*}[t]
\centering
\caption{Parameter quantities for structured lightweight fine-tuning modules in a Transformer block.}
\begin{tabular}{lllll}
\toprule
Modules    & Components    & Variables    & Size    & \#Params. \\
\midrule
           & LN-bias    & $\dot{\beta}$    & $d$    &    \\
BitFit     & MSA-bias    & $\{b_{Q}^{l,h}\}_{h=1}^{H}$    & $\{\frac{d}{H}\} \times H$    &     \\
           &            & $\{b_{K}^{l,h}\}_{h=1}^{H}$    & $\{\frac{d}{H}\} \times H$    &    \\
           &            & $\{b_{V}^{l,h}\}_{h=1}^{H}$    & $\{\frac{d}{H}\} \times H$    & $11d$ \\
           &            & $b_{O}^{l}$    & $d$    &    \\
           & LN-bias    & $\dot{\beta}$    & $d$    &    \\
           & FFN-bias    & $b_{1}^{l}$    & $4d$    &    \\
           &            & $b_{2}^{l}$    & $d$    &    \\
\midrule
VPT        & Prompts    & $P^{l}$    & $p \times d$    & $pd$    \\
\midrule
Adapter    & LN    & $\dot{\gamma}, \dot{\beta}$    & $d, d$    &    \\
           & Projection    & $\mathbf{W}_{\text{down}}, b_{\text{down}}$    & $d \times r, r$    & $(2r+3)d + r$    \\
           &            & $\mathbf{W}_{\text{up}}, b_{\text{up}}$    & $r \times d, d$    &    \\
\midrule
LoRA       & Projection    & $\mathbf{W}_{\text{down}}, \mathbf{W}_{\text{up}}$ (for $\mathbf{W}_{Q}$)    & $d \times r, r \times d$    & $4rd$    \\
           &            & $\mathbf{W}_{\text{down}}, \mathbf{W}_{\text{up}}$ (for $\mathbf{W}_{V}$)    & $d \times r, r \times d$    &    \\
\midrule
AdaptFormer & LN    & $\dot{\gamma}, \dot{\beta}$    & $d, d$    &    \\

            & Projection    & $\mathbf{W}_{\text{down}}, b_{\text{down}}$    & $d \times r, r$    & $(2r+3)d + r + 1$    \\
           &            & $\mathbf{W}_{\text{up}}, b_{\text{up}}$    & $r \times d, d$    &    \\
           & Scaling    & $s$    & $1$    &    \\
\midrule
SG-Adapter & LN    & $\dot{\gamma}, \dot{\beta}$    & $d, d$    &    \\

            & Projection   & ${\mathbf{W}}^{vt}_{\text{proj}}, b^{vt}_{\text{proj}}$    & $d \times d, d$    &    \\
            &            & $\{\mathbf{W}^{x}_{\text{down}}, b^{x}_{\text{down}}\}_{x \in \{v, vt, t\}}$    & $\{d \times r, r\}\times 3$    & $(5r+d+4)d + 3r + 2$    \\
           &            & $\mathbf{W}^{vt}_{\text{up}}, b^{vt}_{\text{up}}$    & $2r \times d, d$    &    \\
           & Scaling    & $s^{vt}, s$    & $1\times 2$    &    \\
\bottomrule
\end{tabular}
\label{tab:ap_param}
\end{table*}
By extending Eq.~(\ref{eq:ap_psx_ptx_5}) and Eq.~(\ref{eq:ap_psx_ptx_6}) to all the classes, for a given sample $\mathbf{x}$ belonging to class $1$, we derive the following formulation:
\begin{equation}
\label{eq:ap_psx_ptx_7}
\begin{split}
P(\mathbf{x})
&= \left(\tau_1 + \frac{P(2)}{P(1)}\cdot\tau_2 + \ldots + \frac{P(C)}{P(1)}\cdot\tau_C \right) \cdot P(1) \\
&\approx \dot{\tau}_1 \cdot P(1).
\end{split}
\end{equation}
Here, $\dot{\tau}_i=\left(\frac{P(1)}{P(i)} \tau_1 + \frac{P(2)}{P(i)}\cdot\tau_2 + \ldots + \frac{P(C)}{P(i)}\cdot\tau_C \right)$ denotes a constant, where $\tau_i\approx \frac{1}{\sqrt{(2\pi)^d |\mathbf{\Sigma}_i|}}\cdot \frac{1}{e^{\frac{1}{2} (\mathbf{x} - \mathbf{m}_i)^{\top} \mathbf{\Sigma}_i^{-1} (\mathbf{x} - \mathbf{m}_i)}}$. Similarly, it can be derived that $P(\mathbf{x})\approx \dot{\tau}_2 \cdot P(2)$ when the sample $\mathbf{x}$ belongs to class $2$. Therefore, for long-tailed training data, the following mathematical relationship holds:
\begin{equation}
\label{eq:ap_psx_ptx_8}
\begin{split}
P(\mathbf{x}_i) \approx \dot{\tau}_i \cdot P(i),
\end{split}
\end{equation}
with $\mathbf{x}_i$ denotes a sample $\mathbf{x}$ from the $i$-th class. However, for the balanced data, where $P(1)=P(2)=\ldots=P(C)=\frac{1}{C}$. Define $\ddot{\tau}_i=\left(\tau_1^{\prime} + \tau_2^{\prime} + \ldots + \tau_C^{\prime} \right)$ and $\tau_i^{\prime}\approx \frac{1}{\sqrt{(2\pi)^d |\mathbf{\Sigma}^{\prime}_i|}}\cdot \frac{1}{e^{\frac{1}{2} (\mathbf{x} - \mathbf{m}^{\prime}_i)^{\top} {\mathbf{\Sigma}^{\prime}}_i^{-1} (\mathbf{x} - \mathbf{m}^{\prime}_i)}}$, with $\mathbf{m}^{\prime}_i$ and ${\mathbf{\Sigma}^{\prime}}_i$ represent the mean and variance for the $i$-th class obtained from balanced data, Eq.~(\ref{eq:ap_psx_ptx_8}) becomes to:
\begin{equation}
\label{eq:ap_psx_ptx_9}
\begin{split}
P(\mathbf{x}_i) \approx \ddot{\tau}_i \cdot \frac{1}{C}.
\end{split}
\end{equation}

Therefore, the marginal probability distribution ratio between long-tailed training data $P^{\prime}_s(\mathbf{x}_i)$ and balanced test data $P_t(\mathbf{x}_i)$ becomes:
\begin{equation}
\label{eq:ap_psx_ptx_10}
\begin{split}
\frac{P^{\prime}_s(\mathbf{x}_i)}{P_t(\mathbf{x}_i)}\approx \frac{\dot{\tau}_i \cdot P(i)}{\ddot{\tau}_i \cdot \frac{1}{C}} = \mu n_i^{\gamma},
\end{split}
\end{equation}
where $P(i)=\frac{n_i}{S_{N}}$, $\mu$ and $\gamma$ are the introduced hyperparameters to control the approximation across the multiple datasets. We also report the \textit{Pearson correlation coefficient}~\citep{kirch2008pearson} and \textit{p-value} between $\{\frac{P^{\prime}_s(\mathbf{x}_i)}{P_t(\mathbf{x}_i)}\}_{i=1}^C$ and the linear mapped sample size of each class on commonly used long-tailed benchmarks, as shown in Fig.~\ref{fig:approximation}. The results consistently indicate a positive correlation between the $\{\frac{P^{\prime}_s(\mathbf{x_i})}{P_t(\mathbf{x_i})}\}_{i=1}^C$ and class-wise sample size, supporting the rationality of $ \frac{P^{\prime}_s(\mathbf{x}_i)}{P_t(\mathbf{x}_i)} \approx \mu n_i^{\gamma}$ for the $i$-th class.

\section{Hyperparameter settings}\label{ap:hyper_setting}
The settings of hyperparameters for all datasets used in our paper are listed in Table~\ref{tab:setting}.

\begin{table*}[!t]
\centering
\setlength{\tabcolsep}{1.5pt}
\caption{Comparison results on CIFAR-100-LT in terms of top-1 accuracy (\%).}
\begin{tabular}{l|cccc|cccc|cccc}
\hline
\multirow{2}{*}{\textbf{Methods}} & 
\multicolumn{4}{c|}{$\beta=100$} & 
\multicolumn{4}{c|}{$\beta=50$} & 
\multicolumn{4}{c}{$\beta=10$} \\ 
& \textbf{All} & \textbf{Head} & \textbf{Med.} & \textbf{Tail} 
& \textbf{All} & \textbf{Head} & \textbf{Med.} & \textbf{Tail} 
& \textbf{All} & \textbf{Head} & \textbf{Med.} & \textbf{Tail}\\ 
\hline
Baseline & \underline{81.9} & 85.3 & \underline{82.5} & \underline{77.2} & 83.2 & 85.3 & 82.6 & 81.5 & 84.9 & 85.9 & 84.1 & 84.7\\
    \ + SG-Adapter & \textbf{82.6} & \underline{85.8} & \textbf{83.0} & \textbf{78.3} & \textbf{84.1} & \textbf{86.3} & \textbf{83.3} & \textbf{82.5} & \textbf{86.2} & \underline{87.2} & \textbf{85.1} & \textbf{86.3}\\
\ + SG-Adapter (w/o ${\mathbf{W}}^{vt}_{\text{proj}}$) & \textbf{82.6} & \textbf{86.2} & \underline{82.5} & \textbf{78.3} & \underline{83.8} & \underline{86.2} & \underline{83.0} & \underline{81.6} & \underline{85.8} & \textbf{87.5} & \underline{84.2} & \underline{85.8}\\
\hline
\end{tabular}
\label{tab:wo_wup_proj_cifar}
\end{table*}
\begin{table*}[!t]
\centering
\setlength{\tabcolsep}{1.5pt}
\caption{Comparison results across large datasets in terms of top-1 accuracy (\%).}
\begin{tabular}{l|cccc|cccc|cccc}
\hline
\multirow{2}{*}{\textbf{Methods}} &
\multicolumn{4}{c|}{\textbf{ImageNet-LT}} & 
\multicolumn{4}{c|}{\textbf{Places-LT}} & 
\multicolumn{4}{c}{\textbf{iNaturalist 2018}} \\ 
& \textbf{All} & \textbf{Head} & \textbf{Med.} & \textbf{Tail} 
& \textbf{All} & \textbf{Head} & \textbf{Med.} & \textbf{Tail} 
& \textbf{All} & \textbf{Head} & \textbf{Med.} & \textbf{Tail}\\ 
\hline
Baseline & \underline{78.4} & \underline{81.3} & \underline{77.4} & \underline{73.5} & 52.2 & \underline{51.6} & 52.9 & 51.7 & \underline{80.6} & \underline{74.3} & \underline{80.5} & 82.3\\
\ + SG-Adapter & \textbf{78.5} & 81.2 & \textbf{77.5} & \textbf{74.0} & \textbf{52.6} & 51.3 & \textbf{53.3} & \textbf{53.2} & \textbf{80.9} & 73.8 & \textbf{80.6} & \textbf{83.0}\\
\ + SG-Adapter (w/o ${\mathbf{W}}^{vt}_{\text{proj}}$) & \underline{78.4} & \textbf{81.6} & 77.3 & 72.8 & \underline{52.5} & \textbf{52.0} & \underline{53.0} & \underline{52.2} & \underline{80.6} & \textbf{74.9} & 80.3 & \underline{82.5}\\
\hline
\end{tabular}
\label{tab:wo_wup_proj_large}
\end{table*}

\begin{table*}[t]
\centering
\setlength{\tabcolsep}{2.5pt}
\caption{Number of learnable parameters introduced across different datasets (M).}
\begin{tabular}{l|cccc}
\hline
\textbf{Methods}
& \textbf{CIFAR-100-LT} 
& \textbf{ImageNet-LT} 
& \textbf{Places-LT} 
& \textbf{iNaturalist 2018}\\ 
\hline
Baseline & 0.10 & 0.62 & 0.18 & 4.75 \\
    \ + SG-Adapter & 0.20 & 1.43 & 0.38 & 11.29\\
\ + SG-Adapter (w/o ${\mathbf{W}}^{vt}_{\text{proj}}$) & 6.96 & 8.19 & 7.13 & 18.04 \\
\hline
\end{tabular}
\label{tab:wo_wup_learn_params}
\end{table*}
\section{Analysis on parameter quantities}\label{sec:ap_param}

We summarize the parameter quantities of fine-tuning modules in Table~\ref{tab:ap_param}, including the proposed multi-modal SG-Adapter and popular lightweight methods following~\citep{shi2023lift}. In our approach, where $d\gg r$, the introduction of the multi-modal up-projection matrix ${\mathbf{W}}^{vt}_{\text{proj}}\in \mathbb{R}^{d\times d}$ leads to a substantial increase in the total number of model parameters. To better understand its impact, we systematically evaluate the importance of integrating ${\mathbf{W}}^{vt}_{\text{proj}}$ into the model by performing a detailed comparison of experiments conducted with and without ${\mathbf{W}}^{vt}_{\text{proj}}$.

\begin{figure}[t]
\centering
\includegraphics[scale=0.27]{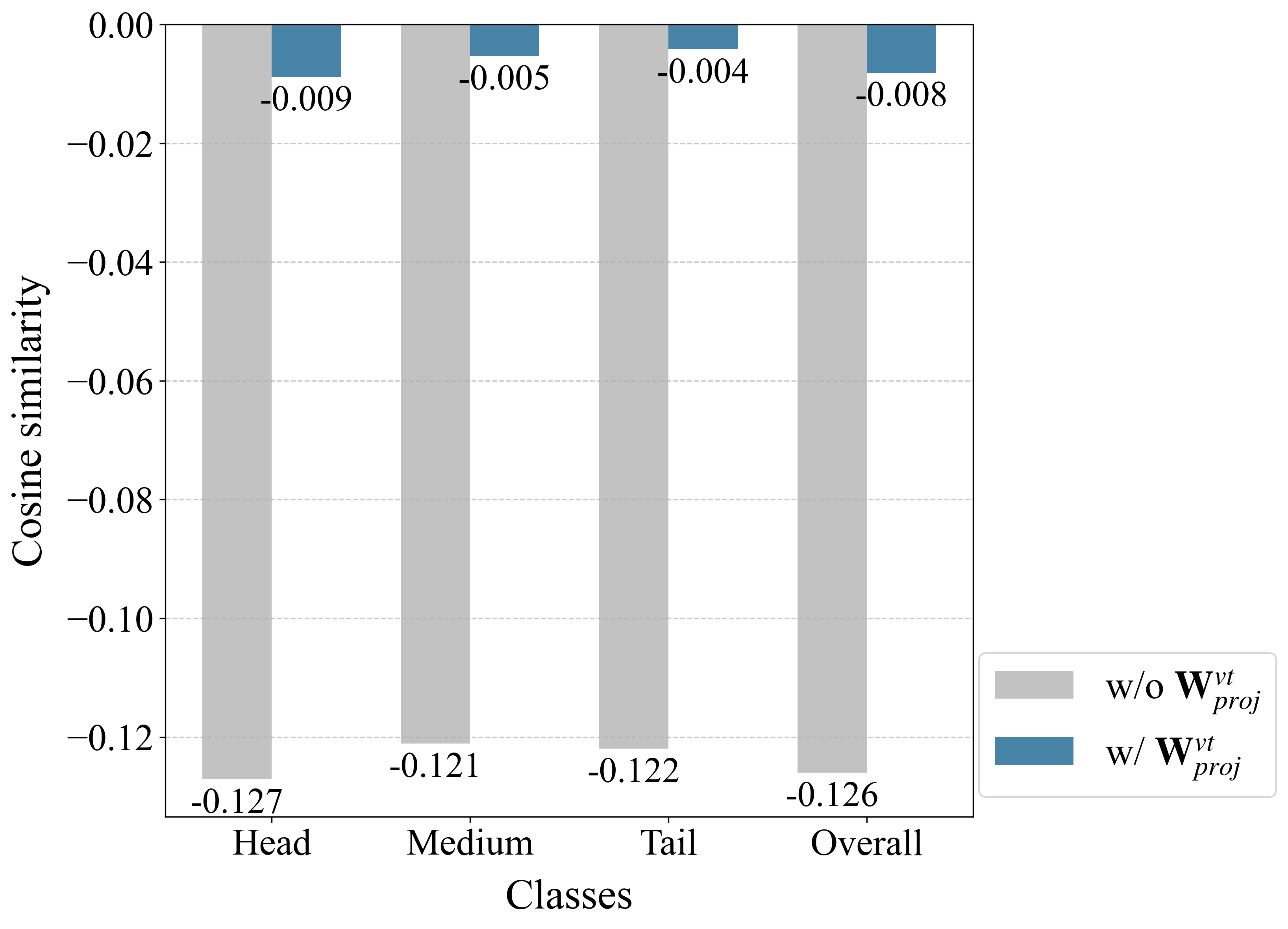}
\caption{Cosine similarity between textual and visual features. `w/o ${\mathbf{W}}^{vt}_{\text{proj}}$' denotes the similarity between plain textual ($\mathbf{f}^{t}$) and visual ($\mathbf{f}^{v}$) features, while `w/ ${\mathbf{W}}^{vt}_{\text{proj}}$' denotes the similarity between projected textual ($\mathbf{f}^{t}\cdot {\mathbf{W}}^{vt}_{\text{proj}}$) and visual features.}
\label{fig:w_up_proj}
\end{figure}

The comparative results of models trained using SG-Adapter with and without ${\mathbf{W}}^{vt}_{\text{proj}}$ are presented in Table~\ref{tab:wo_wup_proj_cifar} and Table~\ref{tab:wo_wup_proj_large}, with the introduced learnable parameters listed in Table~\ref{tab:wo_wup_learn_params}. All models employ the same initialization techniques and are trained with the compensation factor applied consistently. The detailed results on the different classes indicate that it is even more difficult for the model to obtain performance gains on the tail class in the absence of ${\mathbf{W}}^{vt}_{\text{proj}}$. The possible reason is that the multi-modal projection matrix ${\mathbf{W}}^{vt}_{\text{proj}}$ effectively maps the semantics into the embedding space shared with visual features. To validate this, we analyze the cosine similarity between textual and visual features, as presented in Fig.~\ref{fig:w_up_proj}. The results show that, after projection by ${\mathbf{W}}^{vt}_{\text{proj}}$, the textual features exhibit a higher cosine similarity with the visual features. This increased similarity enhances the alignment between the two modalities and facilitates more effective representation learning.

\section{Textual templates list for initialization}\label{ap:templates}
Below we list all 60 textual prompt templates used in our study. In the templates, \texttt{[cls]} represents the placeholder for the class name.

\begin{enumerate}
    \item a photo of a \texttt{[cls]}.
    \item a bad photo of a \texttt{[cls]}.
    \item a photo of many \texttt{[cls]}.
    \item a sculpture of a \texttt{[cls]}.
    \item a photo of the hard to see \texttt{[cls]}.
    \item a low resolution photo of the \texttt{[cls]}.
    \item a rendering of a \texttt{[cls]}.
    \item graffiti of a \texttt{[cls]}.
    \item a bad photo of the \texttt{[cls]}.
    \item a cropped photo of the \texttt{[cls]}.
    \item a tattoo of a \texttt{[cls]}.
    \item the embroidered \texttt{[cls]}.
    \item a photo of a hard to see \texttt{[cls]}.
    \item a bright photo of a \texttt{[cls]}.
    \item a photo of a clean \texttt{[cls]}.
    \item a photo of a dirty \texttt{[cls]}.
    \item a dark photo of the \texttt{[cls]}.
    \item a drawing of a \texttt{[cls]}.
    \item a photo of my \texttt{[cls]}.
    \item the plastic \texttt{[cls]}.
    \item a photo of the cool \texttt{[cls]}.
    \item a close-up photo of a \texttt{[cls]}.
    \item a black and white photo of the \texttt{[cls]}.
    \item a painting of the \texttt{[cls]}.
    \item a painting of a \texttt{[cls]}.
    \item a pixelated photo of the \texttt{[cls]}.
    \item a sculpture of the \texttt{[cls]}.
    \item a bright photo of the \texttt{[cls]}.
    \item a cropped photo of a \texttt{[cls]}.
    \item a plastic \texttt{[cls]}.
    \item a photo of the dirty \texttt{[cls]}.
    \item a jpeg corrupted photo of a \texttt{[cls]}.
    \item a blurry photo of the \texttt{[cls]}.
    \item a photo of the \texttt{[cls]}.
    \item a good photo of the \texttt{[cls]}.
    \item a rendering of the \texttt{[cls]}.
    \item a \texttt{[cls]} in a video game.
    \item a photo of one \texttt{[cls]}.
    \item a doodle of a \texttt{[cls]}.
    \item a close-up photo of the \texttt{[cls]}.
    \item the origami \texttt{[cls]}.
    \item the \texttt{[cls]} in a video game.
    \item a sketch of a \texttt{[cls]}.
    \item a doodle of the \texttt{[cls]}.
    \item an origami \texttt{[cls]}.
    \item a low resolution photo of a \texttt{[cls]}.
    \item the toy \texttt{[cls]}.
    \item a rendition of the \texttt{[cls]}.
    \item a photo of the clean \texttt{[cls]}.
    \item a photo of a large \texttt{[cls]}.
    \item a rendition of a \texttt{[cls]}.
    \item a photo of a nice \texttt{[cls]}.
    \item a photo of a weird \texttt{[cls]}.
    \item a blurry photo of a \texttt{[cls]}.
    \item a cartoon \texttt{[cls]}.
    \item art of a \texttt{[cls]}.
    \item a sketch of the \texttt{[cls]}.
    \item an embroidered \texttt{[cls]}.
    \item a pixelated photo of a \texttt{[cls]}.
    \item itap of the \texttt{[cls]}.
\end{enumerate}

\end{oldappendices}

\end{document}